\documentclass[a4paper]{article}[12pts]

\usepackage[english]{babel}
\usepackage[utf8x]{inputenc}
\usepackage[T1]{fontenc}

\normalsize

\usepackage[a4paper,top=1in,bottom=1in,left=1in,right=1in,marginparwidth=1in]{geometry}

\usepackage{setspace}
\usepackage{amsmath}
\usepackage{amssymb} 
\usepackage{amsthm}
\usepackage{amsfonts, mathtools}
\usepackage{algorithm,algpseudocode}
\usepackage{bm}
\usepackage{bbm}
\usepackage{comment}
\usepackage{graphicx}
\usepackage{multirow, booktabs}
\usepackage{caption}
\usepackage{subcaption}
\usepackage[colorinlistoftodos]{todonotes}
\usepackage[colorlinks=true, allcolors=blue]{hyperref}
\usepackage{pifont}

\DeclareMathOperator*{\argmin}{arg\,min}

\theoremstyle{plain}
\newtheorem{proposition}{Proposition}
\newtheorem{lemma}{Lemma}

\newtheorem{theorem}{Theorem}

\theoremstyle{definition}
\newtheorem{definition}{Definition}
\newtheorem{example}{Example}

\theoremstyle{remark}

\newcommand{\E}{\mathbb{E}}

\usepackage{natbib}
\usepackage{adjustbox}
\usepackage{lipsum}

\newcommand\blfootnote[1]{%
  \begingroup
  \renewcommand\thefootnote{}\footnote{#1}%
  \addtocounter{footnote}{-1}%
  \endgroup
}

\title{When No-Rejection Learning is Consistent for Regression with Rejection}
\author{Xiaocheng Li$^\dagger$, Shang Liu$^\dagger$, Chunlin Sun$^\ddagger$, Hanzhao Wang$^\dagger$}
\date{\small 
$^\dagger$Imperial College Business School, Imperial College London, (xiaocheng.li, s.liu21, h.wang19)@imperial.ac.uk\\
$^\ddagger$Institute for Computational and Mathematical Engineering, Stanford University, chunlin@stanford.edu}

\begin{document}
\maketitle

\onehalfspacing

\blfootnote{Authors alphabetically ordered.}

\begin{abstract}
Learning with rejection has been a prototypical model for studying the human-AI interaction on prediction tasks. Upon the arrival of a sample instance, the model first uses a \textit{rejector} to decide whether to \textit{accept} and use the AI \textit{predictor} to make a prediction or \textit{reject} and defer the sample to humans. Learning such a model changes the structure of the original loss function and often results in undesirable non-convexity and inconsistency issues. For the classification with rejection problem, several works develop consistent surrogate losses for the joint learning of the predictor and the rejector, while there have been fewer works for the regression counterpart. This paper studies the regression with rejection (RwR) problem and investigates a no-rejection learning strategy that uses all the data to learn the predictor. We first establish the consistency for such a strategy under the \textit{weak realizability} condition. Then for the case without the weak realizability, we show that the excessive risk can also be upper bounded with the sum of two parts: prediction error and calibration error. Lastly, we demonstrate the advantage of such a proposed learning strategy with empirical evidence.
\end{abstract}

\section{Introduction}
\label{sec:intro}

The problem of learning with rejection models the teaming of humans and AI systems in accomplishing a prediction task, and it has received increasing attention in recent years due to the adventure of powerful AI tools and critical applications such as medication and healthcare. Learning with rejection has two components: (i) a predictor that predicts the label or target value $Y$ given a feature $X$ and (ii) a rejector that decides whether to make predictions based on uncertainty/low confidence. Upon the rejection, the sample will be deferred to human experts, and the rejection decision will incur a cost. The rejector can be viewed as a binary classifier that assigns the prediction task to either the predictor or the human, and thus it enables the predictor to focus on predicting only the unrejected samples, ideally samples with high prediction confidence. Such problem is also studied under variants that take a similar setup, known as learning under triage \citep{okati2021differentiable}, learning to defer \citep{mozannar2020consistent,verma2022calibrated,verma2023learning,mozannar2023should}, learning under human assistance \citep{de2020regression,de2021classification}, learning to complete human \citep{wilder2020learning}, and selective prediction \citep{geifman2019selectivenet,jiang2020risk}. 
Learning with rejection can also be categorized into two classes based on the underlying task: (i) regression with rejection and (ii) classification with rejection. In this paper, we mainly focus on the regression with rejection problem while some results also generalize to the classification with rejection problem.

A natural way to tackle the learning problem of prediction with rejection is to modify the loss function to incorporate both the predictor loss and the rejection cost. This will enable joint learning of the predictor and the rejector simultaneously by minimizing the modified loss on the training data. If we single out the learning of the predictor, such an approach will only use part of the samples (those unrejected ones) to train the predictor; this is partially justified by the observation that the prediction performance on the rejected samples will not affect the final performance \citep{de2020regression}. For classification problems with a reject option, this approach is also supported by generalization bounds derived using surrogate loss \citep{cortes2016learning}. However, the fact of using part of the training data to train the predictor is quite counter-intuitive to the common sense that more samples will lead to better training of the prediction model. Specifically, for regression with rejection, learning the predictor based on part of the training data can lead to overfitting on the wrong subset \citep{geifman2019selectivenet} or result in a local optimum that performs significantly worse than the global optimum (See Proposition \ref{prop:subopt}). Therefore in this paper, we explore the condition under which \textit{no-rejection learning} -- that treats prediction with rejection just as a standard ML task and utilizes all the training data -- achieves consistency.

\subsection{Our Contribution}

We first establish the consistency of no-rejection learning under a \textit{weak realizability} condition which requires that the predictor function class covers the conditional expectation function. This result implies that the sub-optimality of no-rejection learning will only arise when the underlying function class is not rich enough. Then we proceed with the analysis without this condition and introduce the \textit{truncated loss}. The truncated loss allows full flexibility for the rejector and thus singles out the learning of the predictor. We show that consistency and surrogate property can be derived based on this truncated loss in an easier manner than the joint learning problem. Consequently, it leads to a generalization bound for the regression with rejection problem. We also discuss the learning of the rejector under two environments: fixed cost and fixed budget. Lastly, we use numerical experiments to complement our theoretical results. 


\subsection{Related Literature}

\textbf{Regression with Rejection}. \cite{zaoui2020regression} characterize the optimal predictor and rejector for the function class of all measurable functions, and propose a nonparametric algorithm to learn them. \cite{geifman2019selectivenet} present a neural network-based algorithm with numerical illustrations. \cite{kang2023surrogate} quantify the prediction uncertainty directly, which can also be applied to address the regression with rejection problem. In comparison to these works, we also consider the cost of rejection, while they only investigate the budgeted setting. \cite{jiang2020risk} aim to minimize the rejection rate under a specified level of loss without considering rejection costs. \cite{shah2022selective} consider both rejection costs and reject budgets, highlighting the challenges of optimizing the predictor and rejector even for training samples, and develop a greedy algorithm to solve the problem approximately.

A concurrent work \citep{cheng2024regression} also considers the regression with rejection problem while our work contributes to the problem against it in three aspects: First, \citet{cheng2024regression} establishes a surrogate property for their proposed surrogate loss, but the proposed loss is nonconvex. The nonconvexity brought by loss truncation is unlike the benign nonconvexity convexity for neural networks commonly acknowledged by the deep learning community, but it can result in severe suboptimality as shown in Proposition \ref{prop:subopt}. However, our advocated no-rejection learning enjoys a nice squared loss (which is convex and easy to optimize). Second, we discuss the calibration of the conditional risk and explicitly characterize the suboptimality brought by learning the rejector in Theorem \ref{thm:gen_decompose}, and this part is completely ignored in previous literature. Third, according to our check, the main results of Theorems 5 and 7 in \citet{cheng2024regression} are incorrect as they missed the critical condition that the pre-specified hypothesis class should contain the Bayes-optimal regressor and rejector to make both theorems valid. This is exactly the weak realizability condition that we first introduce to the general problem of learning with rejection.

\textbf{Classification with Rejection}. While the regression with rejection literature has been largely focused on deriving heuristic algorithms, there have been more theoretical developments for the classification with rejection problem, and the key is to introduce a proper surrogate loss. Various surrogate loss functions have been proposed that share the same optimal solution as the original loss function for binary classification and multi-class classification \citep{bartlett2008classification,yuan2010classification,ramaswamy2018consistent,ni2019calibration,verma2022calibrated, charusaie2022sample, cao2022generalizing,mao2024predictor,mao2024theoretically}. Moreover, \cite{cortes2016learning} establish an explicit relationship between the excess loss of the original loss and the surrogate loss. In parallel to the works on classification with rejection, we give the first surrogate loss and consistency analysis for regression with rejection. Different from classification with rejection, no convex surrogate loss has been proposed for regression with rejection, and our work shows that the original squared loss (no-rejection learning) already serves as a good surrogate loss for learning the regressor.

\section{Problem Setup}
\label{sec:setup}
Consider $n$ data samples $\{(X_i,Y_i)\}_{i=1}^n$ drawn independently from an unknown distribution $\mathcal{P}$. The feature vector is $X_i \in \mathcal{X}\subseteq\mathbb{R}^d$, and the target is $Y_i\in \mathcal{Y}\subseteq\mathbb{R}$. The problem of regression with rejection (\texttt{RwR}) considers the regression problem with a rejection option. Specifically, it consists of two components: (i) a regressor $f:\mathcal{X}\rightarrow\mathcal{Y}$, which predicts the target with the feature. (ii) a rejector $r:\mathcal{X}\rightarrow \{0,1\}$, which decides whether to apply the regressor $f$ (when $r(X)=1$) or to defer the sample to human ($r(X)=0$).

Compared to the standard regression problem, the \texttt{RwR} problem introduces a rejection/deferral option. Rejected samples are typically handled by humans at a fixed cost of $c>0$. Consequently, the \texttt{RwR} loss function is defined as follows
$$l_{\texttt{RwR}}(f, r;(X,Y)) = r(X)\cdot l(f(X),Y) +(1-r(X))\cdot c$$
where $l(\cdot,\cdot)$ represents the standard regression loss, say, the squared loss, $l(\hat{Y},Y) = (\hat{Y}-Y)^2$. The motivation of this loss structure is to encourage the deferral of high-risk samples to humans.

The problem of \texttt{RwR} aims to find a regressor and a rejector that jointly minimize the expected loss
\begin{align}
\label{eqdef:loss}\min\limits_{f\in\mathcal{F},r\in\mathcal{G}} L_{\texttt{RwR}}(f,r) \coloneqq \mathbb{E}\left[l_{\texttt{RwR}}(f, r;(X,Y)) \right]
\end{align}
where the expectation is taken with respect to $(X,Y)\sim \mathcal{P}.$ Here $\mathcal{F}$ and $\mathcal{G}$ denote the sets of candidate regressors and rejectors, respectively.


The following proposition from \citet{zaoui2020regression} characterizes the optimal solution of \eqref{eqdef:loss} when only measurability is imposed on the functions classes of $\mathcal{F}$ and $\mathcal{G}.$ It says that the optimal regressor is the conditional expectation, and the optimal rejector rejects samples with a conditional variance larger than $c$.

\begin{proposition}[\cite{zaoui2020regression}]
\label{prop:opt}
Suppose $\mathcal{F}$ contains all measurable functions  that map  from $\mathcal{X}$ to $\mathcal{Y}$, and $\mathcal{G}$ contains all measurable functions  that map  from $\mathcal{X}$ to $\{0,1\}$. Then the optimal regressor $f^*(X)$ and rejector $r^*(X)$ for \eqref{eqdef:loss} are
    \begin{align*}
        f^*(x)  &= \bar{f}(x) \coloneqq \mathbb{E}[Y|X=x],\\
        r^*(x)
        &=
        \begin{cases}
            1, & \text{if \ } \mathbb{E}[(Y-f^*(X))^2|X=x] \leq c,\\
            0, & \text{otherwise}.
        \end{cases}
    \end{align*}
\end{proposition}

Proposition \ref{prop:opt} characterizes the optimal solution, but it does not provide much insight into the learning procedure (which is based on data samples). In the following sections, we will present several results that shed light on the learning procedure, showing how the optimal solution can be achieved.

\section{Consistency under Realizability}
\label{sec:wr}

In this section, we first show a challenge of the joint learning of a regressor $f$ and a rejector $r$ with an example such that directly minimizing the \texttt{RwR} loss will end up with inconsistency/local optima. For the classification setting, some consistent surrogate loss functions have been designed, but there are no existing results for the regression setting to our knowledge. Here we propose a two-step no-rejection learning procedure: we first ignore the \texttt{RwR} loss structure and treat the problem as an ordinary least squares regression problem, then learn a rejector by calibrating the conditional expected loss of the learned regressor. We establish the consistency of such a two-step learning under a weak realizability condition.

\subsection{Challenge of Joint Learning of \textit{f} and \textit{r}}
\label{sec:challenge}

For the classification with rejection problem, there have been various proposed surrogate losses (for the original nonconvex loss) that enjoy consistency guarantees \citep{bartlett2008classification,yuan2010classification,cortes2016boosting,cortes2016learning,ramaswamy2018consistent,ni2019calibration,mozannar2020consistent,verma2022calibrated,mozannar2023should}. However, it turns out to be much more difficult to construct a surrogate loss for the regression with rejection problem. Without a consistency guarantee, the brute-forced approach of joint learning of $f$ and $r$ or alternatively optimizing $f$ and $r$ can lead to bad local minima. To formally illustrate the point, we define the following optimality conditions.

\begin{definition}[Optimality of joint learning]
\label{def:opt}
For any regression with rejection task on $\mathcal{X} \times \mathcal{Y}$ with a regressor class $\mathcal{F}$ (equipped with some norm $\|\cdot\|_{\mathcal{F}}$) and a rejector class $\mathcal{G}$ (equipped with some norm $\|\cdot\|_{\mathcal{G}}$), we define the following optimality conditions for a pair $(f, r) \in \mathcal{F} \times \mathcal{G}$:
\begin{itemize}
    \item[(a)] $(f, r)$ is \textit{globally optimal} if $L_{\texttt{RwR}}(f,r) \leq L_{\texttt{RwR}}(f^\prime,r^\prime)$ for any $(f^\prime, r^\prime) \in \mathcal{F} \times \mathcal{G}$;
    \item[(b)] $(f, r)$ is locally optimal if there exists $\delta_0 > 0$, s.t. $L_{\texttt{RwR}}(f,r) \leq L_{\texttt{RwR}}(f^\prime,r^\prime)$ for any pair $(f',r')$ that $\|f^\prime - f\|_{\mathcal{F}} \leq \delta_0, \|r^\prime - r\|_{\mathcal{G}} \leq \delta_0$;
    \item[(c)] $(f, r)$ is entry-wise optimal if $L_{\texttt{RwR}}(f,r) \leq L_{\texttt{RwR}}(f,r^\prime)$ for any $r^\prime \in \mathcal{G}$ and $L_{\texttt{RwR}}(f,r) \leq L_{\texttt{RwR}}(f^\prime,r)$ for any $f^\prime\in \mathcal{F}$.
\end{itemize}
\end{definition}

With these optimality notions, we formally state the following result that curves some fundamental obstacles in joint learning $(f, r)$.

\begin{proposition}
\label{prop:subopt}
Consider $\mathcal{X} \times \mathcal{Y} = \mathbb{R}^d \times \mathbb{R}$ and a continuous distribution $\mathcal{P}$ with a well-defined probability density function. Suppose the conditional mean $\mathbb{E}[Y|X=x]$ is well-defined for any $x \in \mathbb{R}^d$, and the conditional variance of $Y$ on $X$ varies across the whole domain so that $\min_x \mathbb{E}[(Y - \mathbb{E}[Y|X=x])^2|X=x] < c < \max_x \mathbb{E}[(Y - \mathbb{E}[Y|X=x])^2|X=x]$. Let $\mathcal{F}$ be the class of all measurable functions from $\mathbb{R}^d$ to $\mathbb{R}$ and $\mathcal{G}$ be the class of all measurable functions from $\mathbb{R}^d$ to $\{0, 1\}$. With the distances $\|f^\prime - f\|_{\mathcal{F}} \coloneqq \sup_{x} |f^\prime(x) - f(x)|$ and $\|r^\prime - r\|_{\mathcal{G}} \coloneqq \mathbb{P}_{X\sim \mathcal{P}_X}(r^\prime(X) \neq r(X))$, we have
\begin{itemize}
    \item[(a)] There exists $(f_0, r_0)$ locally optimal but not globally optimal. In other words, joint learning can be trapped by bad initialization;
    \item[(b)] There exists $(f_1, r_1)$ entry-wise optimal but not globally optimal. In other words, alternative optimization can be trapped by entry-wise optimum.
\end{itemize}
\end{proposition}

The intuitions behind Proposition \ref{prop:subopt} are as follows: for the regression with rejection problem, there is a vital difference between \textit{high-risk} samples and \textit{bad-performed} samples, where the high-risk cases are those $X=x$ samples with a large mean squared error under $f$ and the bad-performed samples are those with a large bias term. For any fixed regressor $f$ and feature $X$, the mean squared error can be decomposed into two terms: squared bias and variance. The bias term measures the prediction quality of the regressor $f$ and it is reducible with the choice of $\mathcal{F}$ covering all the measurable functions, while the variance term is an irreducible term and an intrinsic property of the data distribution. However, those two terms are not separately considered when performing the rejection; the rejector focuses on the sum of these two terms. Thus, when the regressor has the rejector to decide which samples to learn from and refrain from learning on bad-performed samples, the regressor would probably miss the potential improvement on those cases with high bias but low variance. In other words, the difficulty of joint learning lies in the difference between regressor $f$ and rejector $r$: the former needs to evaluate bias and variance separately, while the latter cares for the sum. In the context of regression with rejection, or the more general learning with rejection, this observation leads to the suboptimality of the joint learning procedure which ignores the non-convexity and performs a joint optimization of the loss \eqref{eqdef:loss}.

\subsection{Learning with Weak Realizability}
\label{sec:weak_real}
As is discussed in the previous subsection, the difference between learning the regressor and learning the rejector forms the basic difficulties of the regression with rejection problem. Such an observation implies that the learner needs to deal with each learning separately. In this subsection, we formalize this intuition under a condition called \textit{weak realizability}. We first derive a few asymptotic consistency results under the condition; we will discuss the general cases without the condition in Section \ref{sec:lbwr}.
The weak realizability condition states that the function class $\mathcal{F}$ includes the conditional expectation function (as a function mapping $\mathcal{X}$ to $\mathcal{Y}$).

\begin{definition}[Weak realizability]
\label{def:wr}
We say that the distribution $\mathcal{P}$ and the regressor's function class $\mathcal{F}$ satisfy \textit{weak realizability} if 
$$\bar{f}(x) \coloneqq \E[Y|X=x] \in \mathcal{F}.$$
For the rejector function class $\mathcal{G}$, it satisfies \textit{weak realizability} if $\forall f \in \mathcal{F}$,
\[r_f(x) \coloneqq \mathbbm{1}\{\E[(Y-f(X))^2|X=x] \leq c\} \in \mathcal{G}\]
where we call $r_f$ as the rejector induced by $f$.
\end{definition}

We refer to the condition as \textit{weak} realizability in that it only requires the conditional expectation function (or the indicator function of the conditional risk, respectively) belonging to $\mathcal{F}$ (or $\mathcal{G}$, respectively), but it does not require the existence of a regressor-rejector pair that achieves a zero loss as the standard realizability condition. Under such a condition, the \texttt{RwR} problem exhibits a nice learning structure which also extends to the classification problem (See Appendix \ref{appd:ext}).

\begin{proposition}
\label{prop:weak_real}
If the weak realizability condition holds for $\mathcal{P}$ and $\mathcal{F}$, then 
\begin{align*}
\bar{f}(\cdot) \in\argmin_{f\in\mathcal{F}}L_{\mathtt{RwR}}(f,r)
    \end{align*}
for any measurable rejector function $r(\cdot).$
In addition, if the weak realizability condition holds for $\mathcal{G}$, i.e., $r_f \in \mathcal{G}$ for any $f \in \mathcal{F}$, then for any fixed regressor $f \in \mathcal{F}$, $r_f$ minimizes the expected \texttt{RwR} loss that
\[r_f(\cdot) \in \argmin_{r \in \mathcal{G}} L_{\mathtt{RwR}}(f,r).\]
\end{proposition}

While Proposition \ref{prop:subopt} points out the inconsistency of jointly or alternatively learning the regressor and the rejector, Proposition \ref{prop:weak_real} says that this inconsistency disappears if one only \textit{reject to predict} rather than \textit{reject to learn} those bad-performed cases. As is discussed in Section \ref{sec:challenge}, the major concern of joint learning is due to the possibility that a regressor might never improve on those bad-performed cases after rejection. But if one can consistently learn the conditional mean $\bar{f}(x)$, then it naturally leads to the minimizer of the \texttt{RwR} loss for \textit{any} rejector point-wisely. Such a result implies that one can always learn the regressor first and a corresponding rejector later to obtain consistency. In general, we do not need to bother with the rejector when learning the regressor. Thus we can ignore the \texttt{RwR} loss and treat the problem as a standard regression problem to learn the regressor. Such an intuition is formalized in the following proposition.

\begin{proposition}
\label{prop:consistency}
Any regressor $\{f_n\}_{n=1}^\infty$ that is consistent in probability (or almost surely) w.r.t. the squared loss is also consistent in probability (or almost surely) w.r.t. the \texttt{RwR} loss for any regressor $r \in \mathcal{G}$. That is, if
\[\E[(f_n(X) - Y)^2] \rightarrow \E[(f^*(X) - Y)^2]\]
in probability (or almost surely) as $n \rightarrow \infty$, then
\[L_{\mathtt{RwR}}(f_n, r) \rightarrow  L_{\mathtt{RwR}}(f^*, r)\]
in probability (or almost surely) as $n \rightarrow \infty$ for any $r \in \mathcal{G}$. Moreover, if the constant function $1 \in \mathcal{G}$, then the reverse also holds.
\end{proposition}

The following provides a more concrete example of the consistent assumption in Proposition \ref{prop:consistency}.

\begin{example}
\label{eg:parametrized_consistency}
Consider a parameterized family of functions $\mathcal{F}=\{f_{\theta}: f_{\theta}:\mathcal{X}\rightarrow \mathcal{Y}, \theta\in\Theta\}$ with a compact parameter space $\Theta$. Suppose the function is Lipschitz with respect to $\theta$, i.e., $\left\vert  f_{\theta_1}(X)- f_{\theta_2}(X) \right\vert\leq L\| \theta_1-\theta_2\|_2$ with some $L>0$ for all $\theta_1,\theta_2\in\Theta$ and $X\in\mathcal{X}$. Assume $\mathcal{X}$ and $\mathcal{Y}$ are bounded. 
Let
\begin{equation}
 f_{\theta_n}(\cdot)= \argmin_{f\in\mathcal{F}}\sum\limits_{i=1}^{n} l\left( 
        f(X_i),Y_i
    \right). 
    \label{eqn:ERM}
\end{equation}
Then under weak realizability of $\mathcal{F}$, we have $\forall r$,
\[L_{\mathtt{RwR}}(f_{\theta_n}, r) \rightarrow  L_{\mathtt{RwR}}(f^*, r)\]
a.s. as $n\rightarrow \infty$ with $f^*$ defined in Proposition \ref{prop:opt}.
\end{example}

The analysis of Example \ref{eg:parametrized_consistency} shows that although the function $f_{\theta_{n}}$ is learned under the original loss $l$ (e.g., squared loss), it still works optimally for \texttt{RwR} loss. In other words, under the weak realizability condition, the learning is consistent without taking account of the \texttt{RwR} structure. In fact, this is what we mean by \textit{no-rejection learning}. Specifically, consider the empirical version of the \texttt{RwR} loss
\begin{equation}
\min_{f\in \mathcal{F}, r\in\mathcal{G}} \sum_{i=1}^n r(X_i)\cdot l(f(X_i),Y_i) +(1-r(X_i))\cdot c.
\label{eqn:rwr_empi}
\end{equation}
If we take the perspective of learning the regressor $f$, the above empirical loss \eqref{eqn:rwr_empi} essentially only uses part of the training samples (those where $r(X_i)=1$) to learn $f$. This is quite counter-intuitive to the common sense that more training samples will lead to a better model. Proposition \ref{prop:weak_real} with Example \ref{eg:parametrized_consistency} points out that no-rejection learning \eqref{eqn:ERM} which considers all the training samples and simply treats it as a standard regression task, is consistent when the underlying function class $\mathcal{F}$ is rich enough. Moreover, for classification with rejection, while the existing works propose surrogate losses to convexify \eqref{eqn:rwr_empi}, our result says that such design is only necessary when the underlying function class is not rich enough to include the Bayes optimal classifier.

\begin{figure*}[ht!]
     \centering
     \begin{subfigure}[b]{0.32\textwidth}
         \centering
         \includegraphics[width=\textwidth]{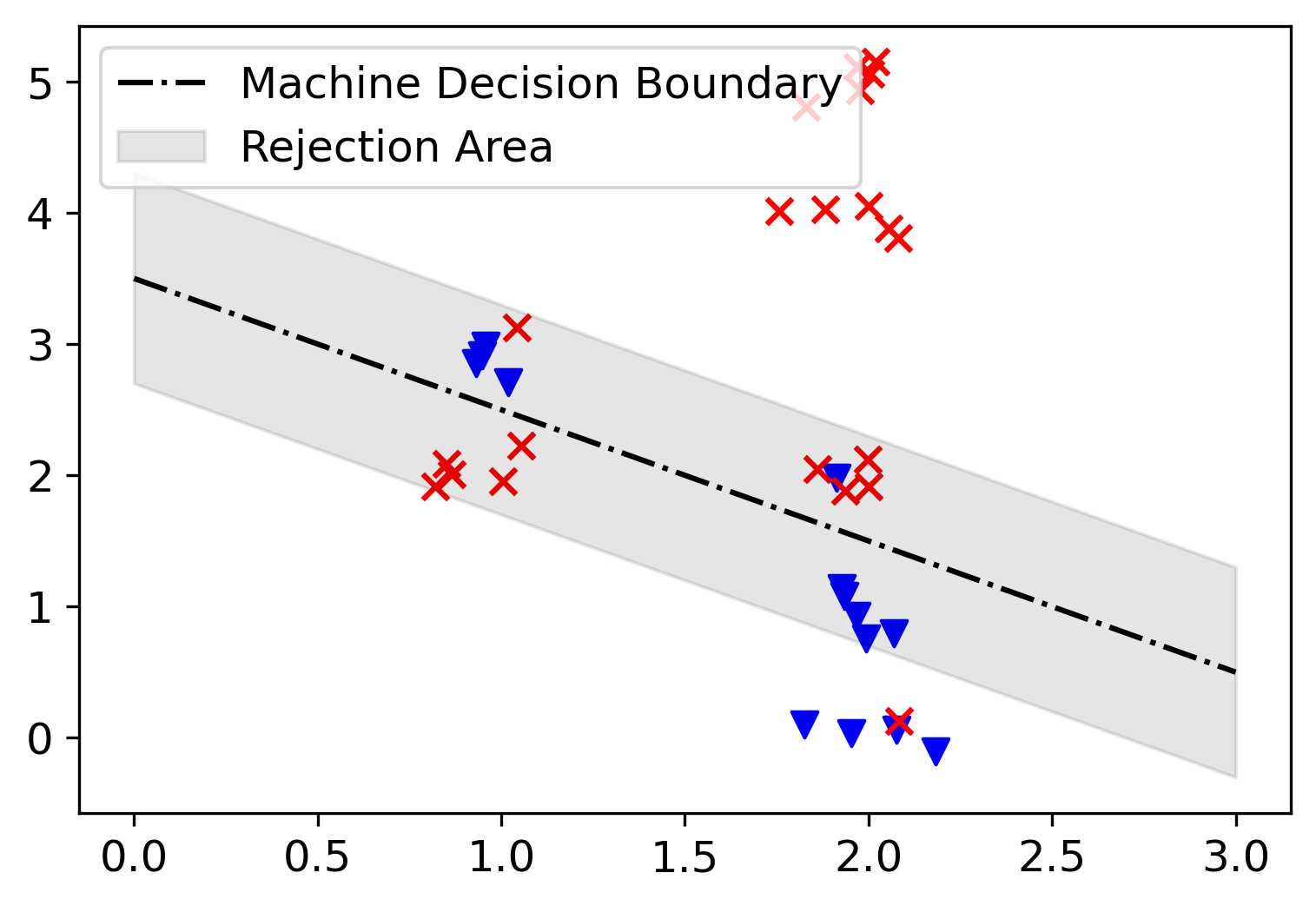}
         \caption{Linear SVM Model 1.}
         \label{fig:mot_exp1}
     \end{subfigure}
     \begin{subfigure}[b]{0.32\textwidth}
         \centering
         \includegraphics[width=\textwidth]{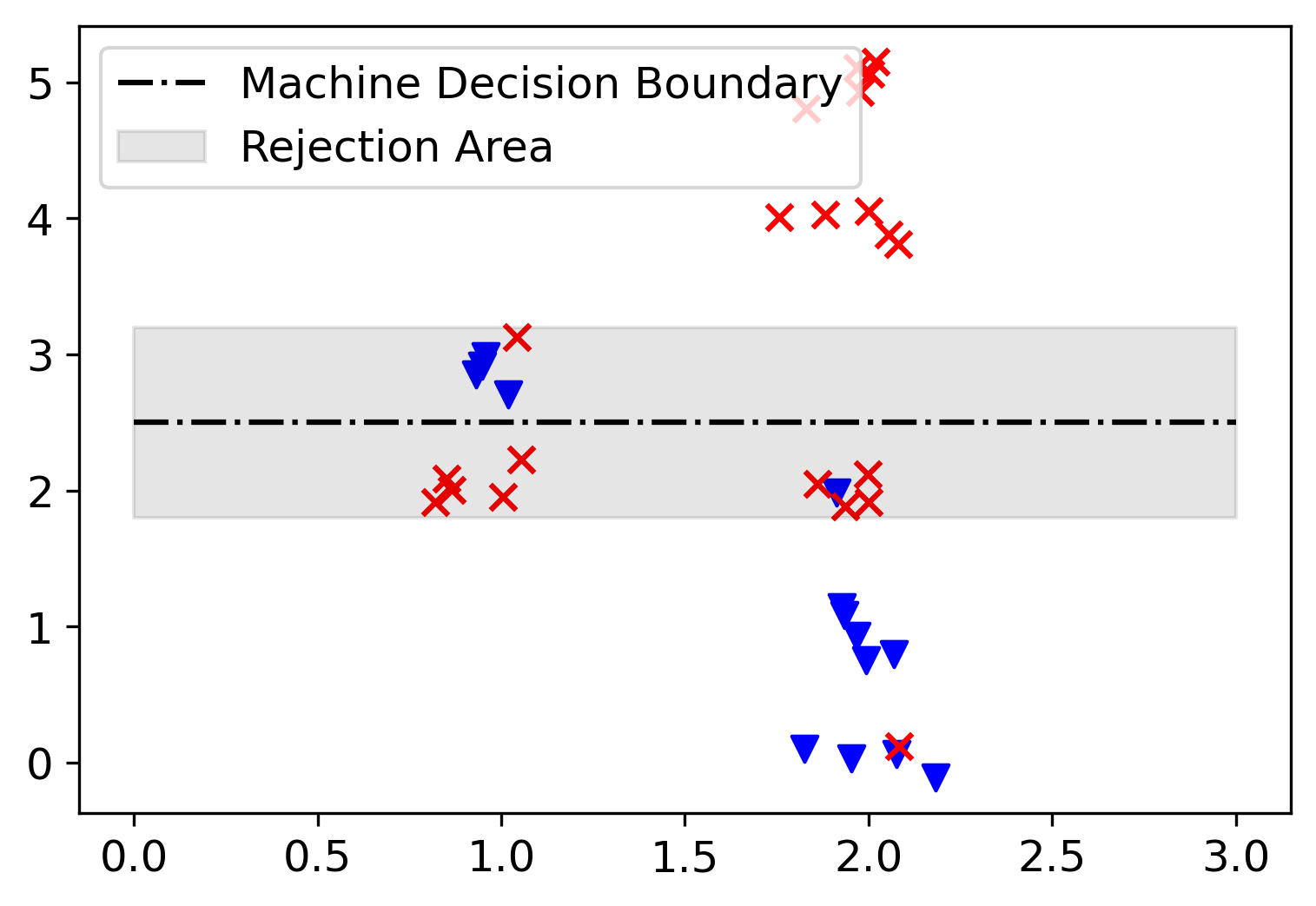}
         \caption{Linear SVM Model 2.}
         \label{fig:mot_exp2}
     \end{subfigure}
     \begin{subfigure}[b]{0.325\textwidth}
         \centering
         \includegraphics[width=\textwidth]{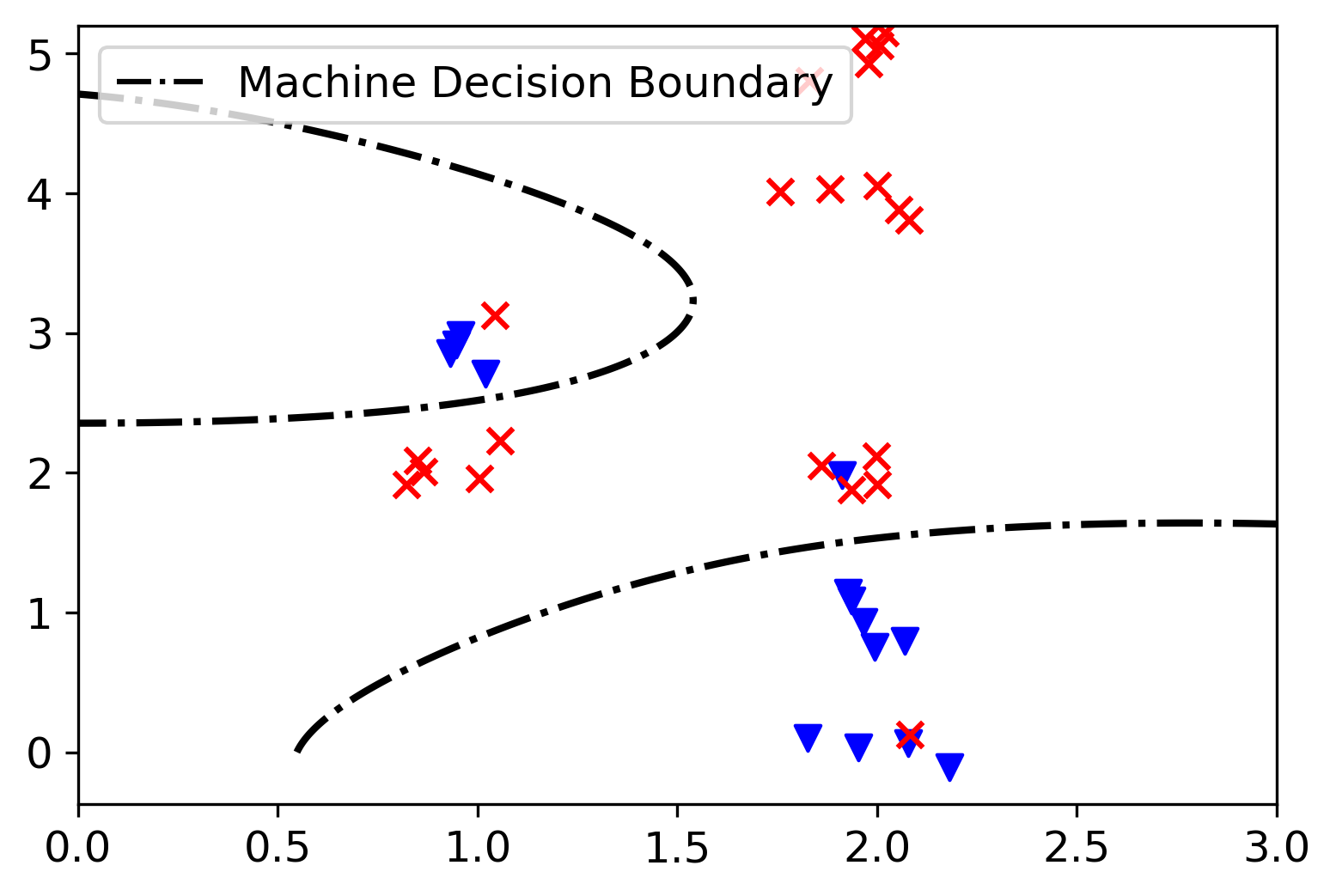}
         \caption{SVM with RBF kernel.}
         \label{fig:mot_exp3}
     \end{subfigure}
\caption{
\small An illustration adapted from \citet{bansal2021most} on classification with rejection. (a) plots the linear SVM model that gives optimal classification accuracy and (b) plots the optimal linear SVM model that optimizes the loss of classification with rejection. That is, (a) ignores the rejection structure and treats the problem as a standard classification problem, while (b) is optimized for the classification with rejection objective. As a consequence, (a) achieves an accuracy of $24/35$, but it needs to reject $20$ points to achieve $34/35$ overall accuracy (assuming the human classifier always correctly predicts). (b) achieves an accuracy of $21/35$, but it only needs to reject $15$ points to achieve $34/35$ overall accuracy. \citet{bansal2021most} uses the paradox to emphasize the importance of accounting for the rejection structure when learning the classifier. In (c), we explain this paradox by the richness of the classifier function class. Specifically, we plot an SVM classifier with a Gaussian radial basis function (RBF) kernel that is learned as the standard classification problem; and this classifier is provably optimal for any measurable rejector (in the sense of Proposition \ref{prop:weak_real}). The contrast between (a) and (b) is a result of the limitation of the classifier class. For the linear SVM model in this example, when it performs well on some region of the data, it will sacrifice the other region. When the function class becomes richer to (in practice, approximately) cover the Bayes optimal classifier, this phenomenon will not exist anymore.}
\label{fig:moti_exp}
\end{figure*}

\cite{bansal2021most} consider classification with rejection and use examples to demonstrate that it will result in a suboptimal predictor/classifier if one ignores the rejector structure and simply performs the standard no-rejection learning. On one hand, the findings highlight the special structure of learning with rejection. On the other hand, our results above say that such suboptimality of the learned predictor/classifier (under no-rejection learning) can be mitigated by the adoption of a richer family of functions $\mathcal{F}$. We further illustrate this intuition in Figure \ref{fig:moti_exp}.

\cite{zaoui2020regression} consider the k-nearest neighbor (k-NN) method for the \texttt{RwR} problem. Our results above generalize and justify their choice of k-NN. Specifically, for k-NN or a nonparametric method in general, one usually imposes an assumption on the Lipschitzness/smoothness of the conditional expectation function $\bar{f}$. This is in fact a special case of our weak realizability condition where the Lipschitzness/smoothness assumption ensures the function class $\mathcal{F}$ of the nonparametric estimators is rich enough to cover the true $\bar{f}$. Hence in their algorithm development, they use all the training samples to learn the k-NN regressor. One goal of our work is to justify such a no-rejection learning procedure with and without the presence of the weak realizability condition, and more importantly, beyond the scope of nonparametric methods.

\section{Learning beyond Weak Realizability}
\label{sec:lbwr}

Now we extend the results in the previous section to the case when weak realizability does not hold. The aim is to establish the squared loss as a surrogate loss of the \texttt{RwR} loss and derive error bounds for no-rejection learning. We also give a simple consistent algorithm to calibrate the conditional risk in order to practically learn the rejector.

\subsection{Surrogate Property}
\label{sec:trun}

We first define the truncated loss as follows
\begin{align}
    \label{eqdef:trun_loss}
    \tilde{L}(f)=\mathbb{E}\left[
            \mathbb{E}\left[ l(f(X),Y)\big \vert X \right] \land c
        \right],
\end{align}
where $a\land b=\min\{a,b\}$ for $a,b\in\mathbb{R}$. Here the inner expectation is taken with respect to the conditional distribution $Y|X$ and the outer expectation is taken with respect to the marginal distribution $X$. Basically, the truncated loss truncates the expected loss given $X$ if it exceeds the threshold $c$.

It is natural to consider the rejector induced by a calibrator $\hat{R}(f, x)$, where the calibrator estimates the conditional risk $R(f, x)\coloneqq \E_Y[(f(X) - Y)^2|X=x]$. We formally define the induced rejector by
\[r_{\hat{R}}(x) \coloneqq \mathbbm{1}\{\hat{R}(f, x) \leq c\}.\]
We shall see in the next proposition that the truncated loss and the original \texttt{RwR} loss are closely related with respect to the induced rejector.

\begin{proposition}
    \label{prop:trun}For the truncated loss, we have: 
    \begin{itemize}
        \item[(a)] For any regressor $f$ and any rejector $r$,
     \[  \tilde{L}(f) \le L_{\mathtt{RwR}}(f,r).\]
        Moreover, the equality holds if $r = r_f$.
        \item[(b)]
        Suppose we have a calibrator $\hat{R}$ to estimate the regressor's conditional risk. Then
        \begin{equation}
        L_{\mathtt{RwR}}(f, r_{\hat{R}}) \leq \tilde{L}(f) + \E\big[\vert\hat{R}(f, X) - R(f, X)\vert\big].\label{eqn:lower_bound_trunc}
        \end{equation}
    \end{itemize}
\end{proposition}

Proposition \ref{prop:trun} establishes the truncated loss as a proxy of the \texttt{RwR} loss and states the relationship between these two. The gap between these two losses can be bounded by a term which we called \textit{calibration error} (the last term in \eqref{eqn:lower_bound_trunc}). We suspend (for the moment) the questions of providing a calibration algorithm to deal with the calibration error. The truncated loss $\tilde{L}$ indeed provides convenience for the algorithm design and analysis in that it singles out the regressor learning problem. Moreover, learning against such a truncated loss is justifiable as the truncated loss can appear in both the upper and lower bounds of the \texttt{RwR} loss in Proposition \ref{prop:trun}. Yet, the non-convexity issue still persists for the truncated loss just as the original \texttt{RwR} loss. Fortunately, the simple squared loss is a provable surrogate loss for the truncated loss and this paves the way for the theoretical development without the previous weak realizability condition.

Define the squared loss as
\[
    L_2(f) \coloneqq \mathbb{E}\left[\left(f(X)-Y\right)^2\right].
\]

The following proposition establishes the squared loss as a \textit{surrogate} loss of the truncated loss, following the analysis of \citet{bartlett2006convexity}.
\begin{proposition}
For any measurable function $f$,  we have      \begin{align}
            \label{ieq:cali_func}
            \tilde{L}(f)-\tilde{L}^*
            \leq
            L_2(f)-L_2^*
        \end{align}
where $\tilde{L}^* \coloneqq \min_{f\in\mathcal{F}} \tilde{L}(f),\ L_2^* \coloneqq \min_{f\in\mathcal{F}} L_2(f)$
with $\mathcal{F}$ here being the class of all measurable functions.
\label{prop:cali}
\end{proposition}

Proposition \ref{prop:cali} upper bounds the excess loss under the truncated loss $\tilde{L}$ by that under the squared loss $L_2.$ Hence learning and optimization against the squared loss $L_2$ have an aligned objective with the truncated loss $\tilde{L}$: the performance guarantee under the former is directly transferable to that under the latter. Note that the squared loss indeed utilizes all the samples and corresponds to no-rejection learning. Thus we advocate no-rejection learning in the following sense: it essentially aims for the squared loss $L_2$, which is a provable surrogate of the truncated loss; the truncated loss subsequently is a proxy of the \texttt{RwR} loss with the gap controlled by Proposition \ref{prop:trun}. We remark that all these arguments are made without the weak realizability condition. In the following, we further pursue the path and establish error bounds for no-rejection learning. Importantly, the error bounds are expressed by the standard estimation error term and the approximation error term, which, we believe, are more tangible than the inevitable gap between local minima and global minimum in joint learning of $f$ and $r$.

\subsection{Error Bounds for No-Rejection Learning}
\label{sec:bdd}

We have the following theorem on the error bound under the \texttt{RwR} loss.

\begin{theorem}
\label{thm:gen_decompose}
For any regressor class $\mathcal{F}$ and calibrator $\hat{R}$ to predict the conditional risk $R(f, x) = \E[(f(X) - Y)^2|X=x]$,
    \begin{align}
        \label{ieq:gen_dis}
\phantom{=}L_{\mathtt{RwR}}(\hat{f}, r_{\hat{R}}) - L_{\mathtt{RwR}}^* &\leq \underbrace{ \mathbb{E}\Big[\big(\hat{f}(X) - f^*(X)\big)^2\Big] }_{\text{prediction error}} \nonumber\\ 
        &+\underbrace{\E\Big[\big\vert \hat{R}(\hat{f}, X) - R(\hat{f}, X)\big\vert\Big]}_{\text{calibration error}}
    \end{align}
where $L_{\mathtt{RwR}}^*=L_{\mathtt{RwR}}(f^*,r^*)$ with $f^*$ and $r^*$ given in Proposition \ref{prop:opt}.
\end{theorem}


Theorem \ref{thm:gen_decompose} upper bounds the \texttt{RwR} error of no-rejection learning $\hat{f}_n$ with two terms: prediction error for the regressor $\hat{f} \in \mathcal{F}$, and calibration error of the conditional risk's calibrator $\hat{R}$. The first term can be analyzed via classical statistical learning theory, so we make a few additional explanations for the second term. How to get a statistically consistent and computationally tractable calibrator with provable guarantee is in general a vast land of research that is of independent interest. Hence we do not bother to theoretically analyze different methods' calibration error bounds explicitly but keep the term to stay as is. The calibration task itself is also another regression task, which means that the calibration error can be further decomposed in a similar way as statistical learning theory does. As previously discussed, one needs weak realizability to get rid of the approximation error completely, which motivates the usage of nonparametric methods for the rejector.

To the best of our knowledge, Theorem \ref{thm:gen_decompose} provides the first generalization bound for the regression with rejection problem. Technically, if we compare our analysis in terms of surrogate loss and generalization bound against the analyses of the classification with rejection problem such as \citet{cortes2016learning}, there are several differences. First, the introduction of $\tilde{L}$ singles out the regressor and makes the analysis much more convenient than the surrogate loss derivation of the joint learning of both the classifier and the rejector. While it seems much more challenging to derive a joint surrogate loss for the regression problem, it does not hurt to follow this ``single-out'' approach, theoretically and algorithmically. Second, the truncated loss for the classification problem truncates the binary loss which is non-convex, while the truncated loss for the regression problem truncates the squared loss which is convex. For the non-convex binary loss, the derivation of one more layer of surrogate loss is inevitable and thus raises the question of whether one can derive a surrogate loss for the classifier and rejector jointly in one shot. But for regression with rejection, the original squared loss provides a natural surrogate loss itself. Thirdly, our result endows no-rejection learning from a practical viewpoint, which is well-motivated for modern machine learning. While the theoretical understanding of the joint learning for linear predictor classes such as linear regression and linear SVM can be of theoretical interest, neural network models have the capacity to meet the weak realizability condition.

\section{Learning the Rejector}
\label{sec:cali}

\subsection{Fixed Cost}

In the previous sections, we mainly focus on the learning of the regressor $f$. Given a learned regressor $\hat{f}_n$, the learning of the rejector is closely related to uncertainty quantification/calibration for regression models. In particular, this can be seen from the loss function $l_{\texttt{RwR}}$ and Proposition \ref{prop:opt}. While the existing literature on calibrating regression models mainly aims to predict quantiles or the conditional distribution of $Y|X$, the goal of the rejector essentially requires a prediction of conditional expected loss $\E[(Y-\hat{f}_n(X))^2|X=x]$ as a function of $x.$ Here we present a nonparametric method that calibrates the conditional expected loss and works as a rejector. The method is by no means the optimal one for all problem settings, but it is simple to implement and generally compatible with any regressor.

\begin{algorithm}[ht!]
\caption{Regressor-agnostic rejector learning}
\label{alg:Calibrate}
\begin{algorithmic}[1]
\State \textbf{Input}: regressor $\hat{f}$, validation data $\mathcal{D}_{\text{val}}=\{(X_i,Y_i)\}_{i=1}^m$, kernel $k(\cdot,\cdot)$.
\State \textcolor{blue}{\%\% Loss calibration}
\State Calculate the empirical loss of each sample in the validation data $\mathcal{D}_{\text{val}}$:
\begin{equation*}
l_i(\hat{f}) \coloneqq l(\hat{f}(X_i),Y_i),\quad  1,...,m.
\end{equation*}
\State Estimate the loss $x$ through kernel estimation:
\begin{equation}
    \label{eq:non_para_alg}
    \hat{R}(\hat{f}, x)\coloneqq \frac{\sum_{i=1}^{m}k(x,X_i)\cdot l_i(\hat{f})}{\sum_{i=1}^{m}k(x,X_i)}.
\end{equation}
\State \textcolor{blue}{\%\% Output rejector}
\State Output the rejector with the given deferral cost $c$:
\begin{align*}
            \hat{r}(x)\coloneqq 
        \begin{cases}
            1, & \text{if \ } \hat{R}(\hat{f}, x) \leq c,\\
            0, & \text{otherwise}.
        \end{cases}
\end{align*}
\State \textbf{Output}: rejector $\hat{r}(x)$.
\end{algorithmic}
\end{algorithm}

Algorithm \ref{alg:Calibrate} learns the rejector based on a given regressor $\hat{f}$. It first uses a nonparametric approach to estimate the condition expected loss $\E[(Y-\hat{f}(X))^2|X=x]$ at a new data point. Then, based on a thresholding rule, it assigns the point to either the regressor or the human. We make several remarks on the algorithm. First, the algorithm is agnostic of the underlying regressor $\hat{f}.$ Second, it utilizes an independent validation dataset different from the one that trains $\hat{f}.$ Third, it can be viewed as a generalization of the k-NN rejector in \citet{zaoui2020regression} which is originally designed to ensure a certain rejection rate. For the theoretical guarantee of the learned rejector, we refer to the approach by \citet{gyorfi2002distribution, liu2023distribution} which can be applied to the context here.

\subsection{Fixed Budget}
Here we consider a different setting of \texttt{RwR} with a fixed rejection budget/ratio rather than a fixed rejection cost. It is formalized as follows: only at most a fixed ratio $\gamma \in (0, 1)$ of the test samples can be deferred to humans, meaning that we should learn a rejector with 
\begin{equation}
\mathbb{P}(r(X) = 1) \geq 1- \gamma. \label{eqn:fixed_ratio}
\end{equation}
In the fixed cost case, the optimal rule is straightforward: defer the sample to humans if and only if its conditional risk is greater than $c$. But in the case without an explicit cost of $c$, one should manually set up a rule to fulfill the requirement \eqref{eqn:fixed_ratio}. For any regressor $f$, the oracle rule is to explicitly derive the threshold $c^*$ such that
\[c^* \coloneqq \sup\{c \in \mathbb{R}: \mathbb{P}_X(\E_Y[(f(X) -  Y)^2 | X] > c^*) \leq \gamma\}. \]
But such a threshold is never obtainable for two reasons: we have no access to the conditional risk nor its distribution. A more practical way is to assume access to a score function $s\colon \mathcal{F} \times \mathcal{X} \rightarrow \mathbb{R}$ to represent the conditional risk $R(f, x) = \E[(f(X) - Y)^2 | X=x]$. For instance, one can use the calibrator $\hat{R}$ that estimates the conditional risk in the position of such a score function. Then we can develop a practical thresholding way in the following proposition.

\begin{proposition}
\label{prop:fixed_ratio}
Suppose we have a score function $s\colon \mathcal{F} \times \mathcal{X} \rightarrow \mathbb{R}$ and a regressor $f \in \mathcal{F}$. For $m$ i.i.d. samples $\{X_j\}_{j=1}^m$ that have the same distribution as $X$ and are independent of the regressor $f$, let
\[\hat{c} \coloneqq \text{the } \lceil (1-\gamma)(m+1) \rceil \text{ smallest value of } \{s(f, X_j)\}_{j=1}^m\]
and define
\[r_{s, f, \hat{c}}(x) \coloneqq \mathbbm{1}\{s(f, x) \leq \hat{c}\},\]
then
\[\mathbb{P}(r_{s, f, \hat{c}}(X) = 1) \geq 1 - \gamma.\]
That is, such a rejector satisfies the requirement \eqref{eqn:fixed_ratio}.
\end{proposition}
The proof of Proposition \ref{prop:fixed_ratio} can be easily adapted from the literature on split conformal prediction (for example, see Proposition 1 of \citet{papadopoulos2002inductive}). We make a short note here to emphasize the requirement of independence between those $m$ samples and the regressor $f$. If, for instance, one uses the same set of samples to accomplish both the task of training $f$ and that of evaluating the scores, one would likely get a set of biased (and probably underestimated) scores. To see it, one just needs to think of the relationships between the residuals of the training samples and the errors of the test samples in the case of linear regression. In fact, such independence is crucial for the split conformal prediction methods \citep{vovk2005algorithmic}.

\begin{table*}[ht!]
\centering
 \resizebox{\columnwidth}{!}{%
\begin{tabular}{c|c|cccc|cccc}
\toprule
\multirow{3}{*}{Dataset}&\multirow{3}{*}{Rej. Budget}&\multicolumn{4}{c|}{Machine loss}& \multicolumn{4}{c}{Rejection rate}\\
&&\multicolumn{2}{c}{Joint learning}& \multicolumn{2}{c|}{No-rejection learning}&\multicolumn{2}{c}{Joint learning}& \multicolumn{2}{c}{No-rejection learning}\\ 
&&SelNet($\alpha=0.5$) & SelNet($\alpha=1$)& NN+SelRej &NN+LossRej& SelNet($\alpha=0.5$) & SelNet($\alpha=1$)& NN+SelRej &NN+LossRej  \\
\midrule
\multirow{1}{*}{Concrete}
 & 0.3 & 45.16 (17.88) & 66.04 (19.12) & 35.92 (4.01) & \textbf{33.35} (5.98) & 0.32 (0.07) & 0.28 (0.06) & 0.26 (0.04) & 0.30 (0.03) \\
\multirow{1}{*}{Airfoil}
 & 0.3 & \textbf{4.85} (0.79) & 4.98 (1.01) & 6.12 (0.72) & 6.55 (0.82) & 0.30 (0.06) & 0.30 (0.05) & 0.31 (0.06) & 0.30 (0.04) \\
\multirow{1}{*}{Parkinsons}
 & 0.3 & 12.13 (3.59) & 11.72 (1.83) & 13.56 (1.96) & \textbf{6.94} (0.64) & 0.30 (0.02) & 0.33 (0.02) & 0.29 (0.02) & 0.31 (0.02) \\
 \multirow{1}{*}{Energy}
 & 0.3 & \textbf{1.19} (0.45) & 1.54 (0.50) & 1.42 (0.12) & 1.25 (0.23) & 0.31 (0.06) & 0.31 (0.06) & 0.29 (0.07) & 0.32 (0.05) \\
\bottomrule
\end{tabular}}
\caption{ Machine loss and rejection rate with budget rate $=0.3$. Full results can be found in the Appendix.}
\label{tab:EXP2}
\end{table*}

One interesting insight from the excessive risk decomposition in Theorem \ref{thm:gen_decompose} is that it is independent of the fixed cost $c$. In the fixed budget/ratio case, if we choose the calibrator $\hat{R}$ as our score function, then our rejector still falls into the category of those induced by calibrators, implying that such a risk decomposition in Theorem \ref{thm:gen_decompose} also holds and applies to this fixed budget setting. Moreover, Proposition \ref{prop:fixed_ratio} provides a simple approach to improve the rejection ratio for fixed-budget \text{RwR} algorithms such as SelectiveNet \citep{geifman2019selectivenet}.

\section{Numerical Experiments}
\label{sec:exp}

We use numerical experiments to show the performance of no-rejection learning on $8$ UCI datasets \citep{asuncion2017uci} under two settings: fixed-cost and fixed-budget. We present part of the results here in Figure \ref{fig:Exp1} and Table \ref{tab:EXP2}.  Full results with details about the experiments can be found in Appendix \ref{apnx: exp_sec}. Our code is available at \url{https://github.com/hanzhao-wang/RwR}.

\textbf{Fixed-cost}: In Figure \ref{fig:Exp1}, we first implement a neural network (NN) model with no-rejection learning and then implement Algorithm \ref{alg:Calibrate} (kNNRej) with  several variants: the logistic regression rejector LogRej (to train a logistic classifier to predict whether the conditional loss $\ge c$), the loss-based rejector LossRej (to first train a linear predictor for the conditional loss and build the rejector by checking whether the predicted loss $\ge c$), the SelectiveNet-based rejector SelRej (to train a  selection head from SelectiveNet (SelNet) \citep{geifman2019selectivenet} to decide whether to select/reject). All these variants are trained based on the second last layer of the feature extracted by the predictor NN. Also, we implement several benchmarks: Differentiable Triage (Triage) \citep{okati2021differentiable}, SelNet \citep{geifman2019selectivenet}, and kNN predictor with reject option (kNN) \citep{zaoui2020regression}. The model NN+kNNRej has a uniform advantage over the other algorithms across different datasets, which highlights (i) the effectiveness of no-rejection learning and (ii) the importance of a rich (non-parametric) rejector class. Also, note always rejecting the sample will result in an $\texttt{RwR}$ loss of $c$, only our algorithm uniformly beats this naive baseline across all datasets.

\begin{figure}[ht!]
    \centering
    \includegraphics[scale=0.5]{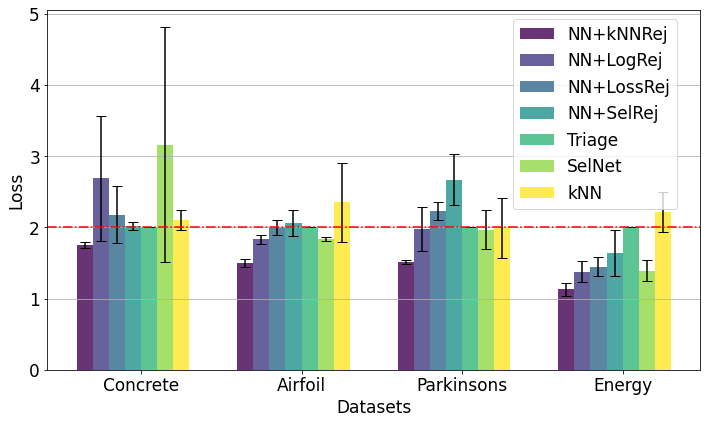}
    \caption{\texttt{RwR} loss with fixed cost $c=2$ (red dashed line).} 
    \label{fig:Exp1}
\end{figure}

\textbf{Fixed-budget}: Table \ref{tab:EXP2} illustrates the performance in terms of the average machine loss and rejection rate on testing data. Four algorithms are examined: SelNet with $\alpha=1$, which jointly learns the regressor and the rejector and the learning of the regressor only focuses on the accepted samples; SelNet with $\alpha=0.5$ which can be viewed as a hybrid of the SelNet with $\alpha=1$ with the no-rejection learning. In other words, the parameter $\alpha$ in SelNet controls the extent to which we focus on no-rejection learning v.s. solely optimizing the accepted samples. Also, we train a neural network model NN with no-rejection learning as regressor with two rejectors, which are both trained based on the second last layer of the feature extracted by the regressor NN: SelRej which uses SelNet's selection head to output rejection score, and LossRej which uses a linear function to predict conditional loss, where both come with a thresholding rejection rule guided by Proposition \ref{prop:fixed_ratio}. We make the following observations. First, the SelNet with $\alpha=0.5$ generally performs better than the SelNet with $\alpha=1$, which indicates the benefit of incorporating no-rejection learning into the objective function. Second, the no-rejection learning algorithms NN+SelRej and NN+LossRej give comparable or even better performance than SelNet. It means that the superior performance achieved by the SelNet in the existing literature should be attributed to its network architecture and the idea of using the high-level feature (second last layer) to train the rejector, rather than the joint learning of regressor and rejector. Third, Proposition \ref{prop:fixed_ratio} and its induced rejector of SelRej and LossRej provide an effective approach to make the rejection rate meet the target level. We defer more experiment results and discussions to Appendix \ref{apnx: exp_sec}.

\section{Conclusion}

In this paper, we study the problem of regression with rejection. To deal with the challenge of non-convexity, we develop a two-step learning procedure, where one first trains a regressor and calibrates the regressor later. In particular, we analyze the property of no-rejection learning which utilizes all the samples in training the regressor. We hope our results provide a starting point for future research on this no-rejection learning perspective of regression with rejection as well as classification with rejection. While our paper mainly focuses on the setting with a fixed deferral cost $c$, we also provide a parallel formulation that considers the regression with the rejection problem with a fixed deferral budget. The fixed deferral cost $c$ gives rise to a new calibration problem that may inspire new calibration algorithms. Establishing more connections between the two settings for both algorithm design and analysis deserves future research.


\bibliographystyle{informs2014} 
\bibliography{main.bib} 

 \appendix
\section{Experiment Details}

\paragraph{Full Experiment Results.}\

For the fixed-cost experiment, Table \ref{tab_apx:EXP1_RwRloss} reports the performance in terms of the averaged \texttt{RwR} loss on testing data with respect to different deferral/rejection costs $c$ and Table \ref{tab_apx:EXP1_Regrate} reports the corresponding rejection rate. We compare Algorithm \ref{alg:Calibrate} (NN+kNNRej) and its variants (NN+LogRej, NN+LossRej, NN+SelRej), against several benchmarks: Differentiable Triage (Triage) \citep{okati2021differentiable}, SelectiveNet (SelNet) \citep{geifman2019selectivenet}, and kNN predictor with reject option (kNN) \citep{zaoui2020regression}. Since all these benchmark algorithms are originally designed for the scenario where the overall rejection rate is capped by a given rate, we modify them for our setting where the deferral cost is positive while no restriction on the rejection rate. In general, the no-rejection learning together with Algorithm \ref{alg:Calibrate} has an advantage over the other algorithms across different datasets and deferral costs. We should note the cost $c$ itself is the baseline benchmark: always rejecting the sample will result in an $\texttt{RwR}$ loss of $c$. However, only our algorithm uniformly outperforms the baseline in all datasets. For the rejection rate on testing data shown in Table \ref{tab_apx:EXP1_Regrate}, we can find the rejection rate is in general decreasing in the deferral cost in our algorithm as expected: there should have fewer samples to be rejected due to the large deferral cost. Another observation is that the Triage algorithm always rejects all samples. Since the Triage algorithm is the only one among implemented algorithms that use only a (potentially small) subset of training data when training the regressor, the trained regressor can be overfitted and thus has large errors in general, which finally causes almost all testing samples to be rejected.

For the fixed-budget experiment, Table \ref{tab:apx_EXP2} illustrates the performance, quantified through average machine loss and rejection rate on testing data, under varying rejection rate budgets. Four algorithms are examined: SelNet with $\alpha=1$, which jointly learns the regressor and the rejector and the learning of the regressor only focuses on the accepted samples; SelNet with $\alpha=0.5$ which can be viewed as a hybrid of the SelNet with $\alpha=1$ with the no-rejection learning; and NN+SelRej and NN+LossRej, two SelNet variants that adhere to a no-rejection learning framework, where NN+SelRej first trains the main body (with no-rejection learning) and then fine-tunes the heads of SelNet, and NN+LossRej follows the same training process except replacing the SelNet's heads with a simple linear layer to predict conditional loss. While all algorithms maintain rejection rates (nearly) within the given budget, there are two notable observations regarding machine loss: (1) SelNet typically outperforms or at least matches SelNet($\alpha=1$); (2) NN+SelRej and NN+LossRej yield losses comparable to those of SelNet. Consequently, it can be inferred that focusing solely on the accepted portion may precipitate overfitting prior to the extraction of high-level features, thereby diminishing performance—a point also discussed in the original work of SelNet. Conversely, employing a no-rejection paradigm in learning the regressor and rejector proves to be at least not harmful to the performance.

Table  \ref{tab_apx:EXP1_RwRloss},  \ref{tab_apx:EXP1_Regrate}, \ref{tab:apx_EXP2} show the full experiments results in Section \ref{sec:exp}. Each test is repeated 10 times and we report the average performance on the test set alongside its standard deviation. For all datasets, the train--validate--test set is split as 70\%--20\%--10\%. 
 
\label{apnx: exp_sec}
\begin{table*}[ht!]
\centering
 \resizebox{\columnwidth}{!}{%
\begin{tabular}{c|c|cccc|ccc}
\toprule
\multirow{2}{*}{Dataset}&\multirow{2}{*}{Cost $c$}&
\multicolumn{4}{c|}{No-rejection learning}& \multicolumn{3}{c}{Benchmark}\\
&&NN+kNNRej&NN+LogRej&NN+LossRej&NN+SelRej&Triage & SelNet &kNN\\
\midrule
\multirow{4}{*}{Concrete}& 0.2 & \textbf{0.19} (0.00) & 0.76 (1.16) & 0.20 (0.00) & 0.20 (0.00) & 0.20 (0.00) & 0.75 (0.68) & 0.20 (0.00) \\
 & 0.5 & \textbf{0.47} (0.01) & 0.78 (0.72) & 0.50 (0.00) & 0.50 (0.00) & 0.50 (0.00) & 1.09 (0.97) & 0.50 (0.00) \\
 & 1.0 & \textbf{0.92} (0.01) & 1.35 (0.41) & 1.01 (0.02) & 1.00 (0.00) & 1.00 (0.00) & 2.82 (2.33) & 1.08 (0.13) \\
 & 2.0 & \textbf{1.75} (0.04) & 2.69 (0.88) & 2.18 (0.40) & 2.02 (0.06) & 2.00 (0.00) & 3.16 (1.65) & 2.11 (0.14) \\
\midrule
\multirow{4}{*}{Wine}
& 0.2 & \textbf{0.10} (0.01) & 0.20 (0.03) & 0.18 (0.02) & 0.24 (0.03) & 0.20 (0.00) & 0.18 (0.03) & 0.27 (0.05) \\
 & 0.5 & \textbf{0.16} (0.01) & 0.25 (0.03) & 0.23 (0.02) & 0.23 (0.04) & 0.50 (0.00) & 0.24 (0.03) & 0.61 (0.09) \\
 & 1.0 & \textbf{0.20} (0.01) & 0.25 (0.03) & 0.24 (0.03) & 0.25 (0.04) & 1.00 (0.00) & 0.25 (0.04) & 0.84 (0.10) \\
 & 2.0 & \textbf{0.23} (0.03) & 0.24 (0.03) & 0.24 (0.03) & 0.24 (0.03) & 2.00 (0.00) & 0.25 (0.04) & 0.95 (0.10) \\
\midrule
\multirow{4}{*}{Airfoil}
& 0.2 & \textbf{0.18} (0.00) & 0.20 (0.01) & 0.28 (0.09) & 0.20 (0.01) & 0.20 (0.00) & 0.21 (0.01) & 0.20 (0.00) \\
 & 0.5 & \textbf{0.44} (0.01) & 0.50 (0.00) & 0.56 (0.05) & 0.51 (0.04) & 0.50 (0.00) & 0.49 (0.00) & 0.50 (0.00) \\
 & 1.0 & \textbf{0.82} (0.02) & 1.01 (0.08) & 1.06 (0.07) & 1.06 (0.08) & 1.00 (0.00) & 0.99 (0.03) & 1.01 (0.03) \\
 & 2.0 & \textbf{1.50} (0.06) & 1.83 (0.06) & 2.00 (0.11) & 2.06 (0.18) & 2.00 (0.00) & 1.84 (0.03) & 2.35 (0.55) \\
\midrule
\multirow{4}{*}{Energy}
 & 0.2 & \textbf{0.17} (0.01) & 0.27 (0.05) & 0.24 (0.05) & 0.21 (0.02) & 0.20 (0.00) & 0.26 (0.10) & 0.21 (0.02) \\
 & 0.5 & \textbf{0.38} (0.02) & 0.51 (0.10) & 0.55 (0.12) & 0.55 (0.08) & 0.50 (0.00) & 0.49 (0.12) & 0.63 (0.11) \\
 & 1.0 & \textbf{0.69} (0.05) & 0.90 (0.10) & 0.86 (0.06) & 1.06 (0.14) & 1.00 (0.00) & 0.86 (0.15) & 1.29 (0.23) \\
 & 2.0 & \textbf{1.13} (0.09) & 1.38 (0.15) & 1.45 (0.13) & 1.64 (0.32) & 2.00 (0.00) & 1.39 (0.15) & 2.21 (0.28) \\
\midrule
\multirow{4}{*}{Housing}
 & 0.2 & \textbf{0.18} (0.01) & 1.59 (0.81) & 1.01 (1.08) & 0.20 (0.00) & 0.20 (0.00) & 1.27 (0.63) & 0.20 (0.00) \\
 & 0.5 & \textbf{0.42} (0.02) & 1.89 (1.34) & 1.57 (1.36) & 0.59 (0.14) & 0.50 (0.00) & 1.31 (0.67) & 0.59 (0.23) \\
 & 1.0 & \textbf{0.79} (0.02) & 3.71 (5.02) & 2.18 (1.52) & 1.56 (0.47) & 1.00 (0.00) & 2.23 (1.16) & 1.15 (0.29) \\
 & 2.0 & \textbf{1.43} (0.05) & 2.80 (0.76) & 3.74 (1.39) & 4.27 (1.29) & 2.00 (0.00) & 2.47 (0.50) & 2.48 (0.51) \\
\midrule
\multirow{4}{*}{Solar}
& 0.2 & \textbf{0.08} (0.01) & 0.22 (0.08) & 0.44 (0.23) & 0.33 (0.19) & 0.20 (0.00) & 0.18 (0.04) & 0.23 (0.14) \\
 & 0.5 & \textbf{0.15} (0.01) & 0.44 (0.15) & 0.50 (0.28) & 0.60 (0.37) & 0.50 (0.00) & 0.45 (0.15) & 0.36 (0.17) \\
 & 1.0 & \textbf{0.22} (0.03) & 0.56 (0.12) & 0.71 (0.36) & 0.68 (0.35) & 1.00 (0.00) & 0.52 (0.20) & 0.60 (0.34) \\
 & 2.0 & \textbf{0.32} (0.04) & 0.72 (0.34) & 0.75 (0.35) & 0.68 (0.33) & 2.00 (0.00) & 0.68 (0.35) & 0.70 (0.36) \\
\midrule
\multirow{4}{*}{Forest}
& 0.2 & \textbf{0.17} (0.01) & 0.68 (0.53) & 0.65 (0.26) & 0.71 (0.81) & 0.20 (0.00) & 0.54 (0.40) & 0.20 (0.00) \\
 & 0.5 & \textbf{0.39} (0.02) & 1.03 (0.32) & 1.12 (0.37) & 1.25 (0.56) & 0.50 (0.00) & 0.78 (0.32) & 0.51 (0.03) \\
 & 1.0 & \textbf{0.68} (0.06) & 1.30 (0.25) & 1.44 (0.20) & 2.17 (0.74) & 1.00 (0.00) & 1.25 (0.42) & 1.22 (0.24) \\
 & 2.0 & \textbf{1.06} (0.10) & 2.19 (0.33) & 2.17 (0.49) & 2.58 (1.08) & 2.00 (0.00) & 2.17 (0.42) & 2.01 (0.32) \\
\midrule
\multirow{4}{*}{Parkinsons}
 & 0.2 & \textbf{0.18} (0.00) & 0.21 (0.04) & 0.38 (0.05) & 0.20 (0.00) & 0.20 (0.00) & 0.20 (0.00) & 0.43 (0.18) \\
 & 0.5 & \textbf{0.44} (0.00) & 0.50 (0.00) & 0.81 (0.15) & 0.51 (0.01) & 0.50 (0.00) & 0.60 (0.23) & 0.69 (0.19) \\
 & 1.0 & \textbf{0.82} (0.01) & 1.06 (0.14) & 1.28 (0.10) & 1.13 (0.09) & 1.00 (0.00) & 1.17 (0.27) & 1.12 (0.27) \\
 & 2.0 & \textbf{1.51} (0.03) & 1.98 (0.31) & 2.23 (0.12) & 2.67 (0.36) & 2.00 (0.00) & 1.97 (0.27) & 1.99 (0.42) \\
\bottomrule
\end{tabular}}
\caption{Full results for Figure \ref{fig:Exp1}. \texttt{RwR} loss on the testing data.}
\label{tab_apx:EXP1_RwRloss}
\end{table*}

\begin{table*}[ht!]
\centering
 \resizebox{\columnwidth}{!}{%
\begin{tabular}{c|c|cccc|ccc}
\toprule
\multirow{2}{*}{Dataset}&\multirow{2}{*}{Cost $c$}&
\multicolumn{4}{c|}{No-rejection learning}& \multicolumn{3}{c}{Benchmark}\\
&&NN+kNNRej&NN+LogRej&NN+LossRej&NN+SelRej&Triage & SelNet &kNN\\
\midrule
\multirow{4}{*}{Concrete}& 0.2 & 0.95 (0.01) & 0.96 (0.05) & 1.00 (0.00) & 1.00 (0.00) & 1.00 (0.00) & 0.95 (0.03) & 1.00 (0.00) \\
 & 0.5 & 0.90 (0.02) & 0.98 (0.03) & 1.00 (0.00) & 1.00 (0.00) & 1.00 (0.00) & 0.95 (0.03) & 1.00 (0.00) \\
 & 1.0 & 0.88 (0.02) & 0.95 (0.05) & 1.00 (0.00) & 1.00 (0.00) & 1.00 (0.00) & 0.90 (0.10) & 0.99 (0.01) \\
 & 2.0 & 0.80 (0.02) & 0.96 (0.04) & 1.00 (0.01) & 1.00 (0.01) & 1.00 (0.00) & 0.93 (0.06) & 0.98 (0.01) \\
\midrule
\multirow{4}{*}{Wine}
& 0.2 & 0.31 (0.04) & 0.59 (0.18) & 0.03 (0.01) & 0.03 (0.01) & 1.00 (0.00) & 0.59 (0.13) & 0.83 (0.03) \\
 & 0.5 & 0.13 (0.02) & 0.00 (0.00) & 0.00 (0.00) & 0.00 (0.00) & 1.00 (0.00) & 0.00 (0.00) & 0.52 (0.05) \\
 & 1.0 & 0.05 (0.02) & 0.00 (0.00) & 0.00 (0.00) & 0.00 (0.00) & 1.00 (0.00) & 0.01 (0.02) & 0.23 (0.03) \\
 & 2.0 & 0.01 (0.01) & 0.00 (0.00) & 0.00 (0.00) & 0.00 (0.00) & 1.00 (0.00) & 0.00 (0.00) & 0.04 (0.02) \\
\midrule
\multirow{4}{*}{Airfoil}
 & 0.2 & 0.87 (0.03) & 0.98 (0.02) & 1.00 (0.00) & 1.00 (0.00) & 1.00 (0.00) & 0.97 (0.01) & 1.00 (0.00) \\
 & 0.5 & 0.81 (0.03) & 0.99 (0.01) & 1.00 (0.01) & 1.00 (0.01) & 1.00 (0.00) & 0.97 (0.02) & 1.00 (0.00) \\
 & 1.0 & 0.74 (0.03) & 0.92 (0.09) & 0.92 (0.04) & 0.92 (0.04) & 1.00 (0.00) & 0.87 (0.07) & 1.00 (0.00) \\
 & 2.0 & 0.64 (0.04) & 0.82 (0.03) & 0.78 (0.04) & 0.78 (0.04) & 1.00 (0.00) & 0.83 (0.03) & 0.98 (0.02) \\
\midrule
\multirow{4}{*}{Energy}
 & 0.2 & 0.77 (0.05) & 0.75 (0.05) & 0.91 (0.04) & 0.91 (0.04) & 1.00 (0.00) & 0.70 (0.09) & 1.00 (0.01) \\
 & 0.5 & 0.64 (0.05) & 0.71 (0.10) & 0.68 (0.06) & 0.68 (0.06) & 1.00 (0.00) & 0.66 (0.09) & 0.93 (0.02) \\
 & 1.0 & 0.54 (0.07) & 0.65 (0.13) & 0.52 (0.05) & 0.52 (0.05) & 1.00 (0.00) & 0.65 (0.06) & 0.70 (0.04) \\
 & 2.0 & 0.40 (0.06) & 0.42 (0.13) & 0.29 (0.06) & 0.29 (0.06) & 1.00 (0.00) & 0.41 (0.12) & 0.39 (0.06) \\
\midrule
\multirow{4}{*}{Housing}
 & 0.2 & 0.87 (0.05) & 0.60 (0.09) & 1.00 (0.00) & 1.00 (0.00) & 1.00 (0.00) & 0.66 (0.03) & 0.99 (0.01) \\
 & 0.5 & 0.78 (0.03) & 0.65 (0.10) & 0.96 (0.03) & 0.96 (0.03) & 1.00 (0.00) & 0.72 (0.07) & 0.99 (0.02) \\
 & 1.0 & 0.69 (0.03) & 0.67 (0.13) & 0.79 (0.08) & 0.79 (0.08) & 1.00 (0.00) & 0.67 (0.07) & 0.97 (0.04) \\
 & 2.0 & 0.57 (0.04) & 0.63 (0.10) & 0.47 (0.10) & 0.47 (0.10) & 1.00 (0.00) & 0.70 (0.08) & 0.92 (0.06) \\
\midrule
\multirow{4}{*}{Solar}
 & 0.2 & 0.28 (0.04) & 0.63 (0.17) & 0.19 (0.04) & 0.19 (0.04) & 1.00 (0.00) & 0.62 (0.13) & 0.35 (0.05) \\
 & 0.5 & 0.20 (0.03) & 0.50 (0.21) & 0.08 (0.04) & 0.08 (0.04) & 1.00 (0.00) & 0.48 (0.20) & 0.24 (0.05) \\
 & 1.0 & 0.11 (0.03) & 0.28 (0.22) & 0.03 (0.02) & 0.03 (0.02) & 1.00 (0.00) & 0.24 (0.16) & 0.16 (0.05) \\
 & 2.0 & 0.07 (0.02) & 0.03 (0.08) & 0.01 (0.01) & 0.01 (0.01) & 1.00 (0.00) & 0.03 (0.05) & 0.07 (0.04) \\
\midrule
\multirow{4}{*}{Forest}
 & 0.2 & 0.77 (0.05) & 0.86 (0.08) & 0.89 (0.07) & 0.89 (0.07) & 1.00 (0.00) & 0.88 (0.08) & 1.00 (0.01) \\
 & 0.5 & 0.66 (0.07) & 0.82 (0.08) & 0.55 (0.07) & 0.55 (0.07) & 1.00 (0.00) & 0.90 (0.06) & 0.97 (0.02) \\
 & 1.0 & 0.51 (0.08) & 0.81 (0.10) & 0.16 (0.05) & 0.16 (0.05) & 1.00 (0.00) & 0.87 (0.11) & 0.74 (0.08) \\
 & 2.0 & 0.31 (0.05) & 0.63 (0.28) & 0.01 (0.01) & 0.01 (0.01) & 1.00 (0.00) & 0.54 (0.38) & 0.38 (0.06) \\
\midrule
\multirow{4}{*}{Parkinsons}
& 0.2 & 0.88 (0.02) & 0.97 (0.05) & 1.00 (0.00) & 1.00 (0.00) & 1.00 (0.00) & 0.99 (0.01) & 0.77 (0.01) \\
 & 0.5 & 0.81 (0.01) & 0.99 (0.00) & 1.00 (0.00) & 1.00 (0.00) & 1.00 (0.00) & 0.94 (0.07) & 0.73 (0.01) \\
 & 1.0 & 0.73 (0.02) & 0.95 (0.07) & 0.96 (0.02) & 0.96 (0.02) & 1.00 (0.00) & 0.90 (0.08) & 0.70 (0.01) \\
 & 2.0 & 0.64 (0.02) & 0.87 (0.08) & 0.80 (0.03) & 0.80 (0.03) & 1.00 (0.00) & 0.85 (0.09) & 0.65 (0.02) \\
 \bottomrule
\end{tabular}}
\caption{Full results for Figure \ref{fig:Exp1}. Rejection rate on the testing data.}
\label{tab_apx:EXP1_Regrate}
\end{table*}

\begin{table*}[ht!]
\centering
 \resizebox{\columnwidth}{!}{%
\begin{tabular}{c|c|cccc|cccc}
\toprule
\multirow{3}{*}{Dataset}&\multirow{3}{*}{Rej. Budget}&\multicolumn{4}{c|}{Machine loss}& \multicolumn{4}{c}{Rejection rate}\\
&&\multicolumn{2}{c}{Joint learning}& \multicolumn{2}{c|}{No-rejection learning}&\multicolumn{2}{c}{Joint learning}& \multicolumn{2}{c}{No-rejection learning}\\ 
&&SelNet($\alpha=0.5$) & SelNet($\alpha=1$)& NN+SelRej &NN+LossRej& SelNet($\alpha=0.5$) & SelNet($\alpha=1$)& NN+SelRej &NN+LossRej  \\
\midrule
\multirow{3}{*}{Concrete}
 & 0.1 & 56.05 (11.67) & 71.28 (17.47) & 44.99 (6.95) & \textbf{38.55} (6.49) & 0.10 (0.04) & 0.14 (0.03) & 0.10 (0.03) & 0.11 (0.02) \\
 & 0.2 & 48.39 (8.84) & 67.03 (15.19) & 41.12 (4.86) & \textbf{35.71} (5.25) & 0.20 (0.04) & 0.20 (0.05) & 0.18 (0.03) & 0.21 (0.04) \\
 & 0.3 & 45.16 (17.88) & 66.04 (19.12) & 35.92 (4.01) & \textbf{33.35} (5.98) & 0.32 (0.07) & 0.28 (0.06) & 0.26 (0.04) & 0.30 (0.03) \\
\midrule
\multirow{3}{*}{Wine}
& 0.1 & 0.23 (0.03) & 0.24 (0.04) & 0.23 (0.03) & \textbf{0.19} (0.03) & 0.05 (0.01) & 0.06 (0.02) & 0.07 (0.02) & 0.15 (0.04) \\
 & 0.2 & 0.22 (0.04) & 0.21 (0.04) & 0.21 (0.04) & \textbf{0.17} (0.03) & 0.18 (0.03) & 0.19 (0.03) & 0.19 (0.03) & 0.28 (0.03) \\
 & 0.3 & 0.20 (0.03) & 0.19 (0.04) & 0.21 (0.04) & \textbf{0.18} (0.04) & 0.26 (0.03) & 0.26 (0.05) & 0.26 (0.04) & 0.39 (0.04) \\
\midrule
\multirow{3}{*}{Airfoil}
 & 0.1 & 8.08 (1.27) & \textbf{8.07} (1.28) & 8.78 (0.88) & 9.01 (1.29) & 0.10 (0.04) & 0.10 (0.04) & 0.10 (0.03) & 0.10 (0.02) \\
 & 0.2 & 6.35 (1.16) & \textbf{5.66} (0.92) & 7.51 (1.34) & 7.26 (0.89) & 0.20 (0.04) & 0.20 (0.05) & 0.19 (0.05) & 0.21 (0.03) \\
 & 0.3 & \textbf{4.85} (0.79) & 4.98 (1.01) & 6.12 (0.72) & 6.55 (0.82) & 0.30 (0.06) & 0.30 (0.05) & 0.31 (0.06) & 0.30 (0.04) \\
\midrule
\multirow{3}{*}{Energy}
& 0.1 & \textbf{2.91} (0.68) & 3.06 (0.62) & 3.30 (0.64) & 3.04 (0.56) & 0.11 (0.04) & 0.09 (0.03) & 0.11 (0.03) & 0.11 (0.04) \\
 & 0.2 & 1.80 (0.48) & 1.92 (0.67) & 1.80 (0.29) & \textbf{1.51} (0.16) & 0.19 (0.04) & 0.20 (0.04) & 0.19 (0.05) & 0.21 (0.04) \\
 & 0.3 & \textbf{1.19} (0.45) & 1.54 (0.50) & 1.42 (0.12) & 1.25 (0.23) & 0.31 (0.06) & 0.31 (0.06) & 0.29 (0.07) & 0.32 (0.05) \\
\midrule
\multirow{3}{*}{Housing}
& 0.1 & \textbf{7.76} (2.42) & 8.11 (2.25) & 7.83 (2.61) & \textbf{7.76} (1.82) & 0.13 (0.05) & 0.12 (0.05) & 0.11 (0.07) & 0.13 (0.04) \\
 & 0.2 & \textbf{6.71} (1.82) & 7.15 (2.57) & 7.06 (2.39) & 7.77 (1.96) & 0.21 (0.04) & 0.23 (0.04) & 0.23 (0.08) & 0.23 (0.04) \\
 & 0.3 & \textbf{6.01} (2.55) & 6.25 (3.01) & 6.03 (1.97) & 6.75 (2.46) & 0.32 (0.05) & 0.31 (0.04) & 0.35 (0.07) & 0.32 (0.06) \\
 \midrule
 \multirow{3}{*}{Solar}
  & 0.1 & 0.60 (0.32) & \textbf{0.58} (0.33) & 0.61 (0.35) & 0.65 (0.41) & 0.08 (0.04) & 0.09 (0.04) & 0.10 (0.05) & 0.11 (0.03) \\
 & 0.2 & 0.41 (0.26) & 0.40 (0.26) & \textbf{0.38} (0.30) & 0.63 (0.42) & 0.18 (0.03) & 0.18 (0.03) & 0.20 (0.03) & 0.20 (0.04) \\
 & 0.3 & 0.38 (0.30) & 0.39 (0.32) & \textbf{0.24} (0.16) & 0.60 (0.34) & 0.30 (0.06) & 0.29 (0.06) & 0.30 (0.07) & 0.30 (0.04) \\
 \midrule
 \multirow{3}{*}{Forest}
 & 0.1 & 2.62 (1.59) & 2.36 (0.92) & 2.54 (1.02) & \textbf{2.31} (0.57) & 0.08 (0.06) & 0.09 (0.06) & 0.07 (0.04) & 0.13 (0.06) \\
 & 0.2 & 2.29 (0.75) & 2.31 (0.79) & \textbf{2.21} (0.51) & 2.31 (0.65) & 0.16 (0.07) & 0.20 (0.08) & 0.21 (0.09) & 0.31 (0.07) \\
 & 0.3 & 2.44 (1.13) & 2.63 (1.40) & 2.30 (1.16) & \textbf{2.22} (0.82) & 0.28 (0.10) & 0.29 (0.08) & 0.32 (0.12) & 0.40 (0.09) \\
  \midrule
\multirow{3}{*}{Parkinsons}
& 0.1 & 13.57 (1.54) & 14.41 (2.03) & 16.17 (2.50) & \textbf{10.83} (1.26) & 0.16 (0.02) & 0.19 (0.02) & 0.15 (0.01) & 0.10 (0.01) \\
 & 0.2 & 13.50 (2.39) & 13.67 (1.18) & 16.10 (2.35) & \textbf{8.21} (0.65) & 0.21 (0.01) & 0.26 (0.02) & 0.21 (0.02) & 0.21 (0.02) \\
 & 0.3 & 12.13 (3.59) & 11.72 (1.83) & 13.56 (1.96) & \textbf{6.94} (0.64) & 0.30 (0.02) & 0.33 (0.02) & 0.29 (0.02) & 0.31 (0.02) \\
\bottomrule
\end{tabular}}
\caption{Extended Table for Table \ref{tab:EXP2}. Machine loss and rejection rate on the testing data.}
\label{tab:apx_EXP2}
\end{table*}

\paragraph{Datasets.}\
\begin{itemize}
\item \textit{Airfoil Self-Noise} (Airfoil) \citep{misc_airfoil_self-noise_291} is collected by NASA to test two and three-dimensional airfoil blade sections conducted in an anechoic wind tunnel. It has 1503 instances with five features. The output is the scaled sound pressure level.
\item \textit{Concrete Compressive Strength} (Concrete) contains 1030 instances with eight features. The output is the compressive strength of concrete, which is a highly non-linear function of its ingredients (seven features) and age (one feature) as indicated in the original data description \citep{misc_concrete_compressive_strength_165}.
\item  \textit{Wine Quality} (Wine) \citep{misc_wine_quality_186} is to predict the wine quality based on physicochemical tests. We use the red wine part in the dataset, which contains 1599 instances with 11 features.
\item \textit{Energy Efficiency} (Energy) \citep{misc_energy_efficiency_242} is to predict the energy efficiency of the buildings as a function of building parameters. It contains 768 instances and 8 features.
\item \textit{Boston Housing} (Housing) is to predict the median value of a home collected in Boston, originally published by \citet{harrison1978hedonic}. It contains 506 instances and 14 features.
\item \textit{Solar Flare} (Solar) \citep{misc_solar_flare_89} is to predict the number of solar flares of a certain class that occur in a 24 hour period. In the experiment, we only predict for X-class solar flares. It contains 1066 instances and 10 features.
\item \textit{Forest Fires} (Forest) \citep{misc_forest_fires_162} is to predict the burned area of forest fires. It contains 517 instances and 12 features.
\item \textit{Parkinsons Telemonitoring} (Parkinsons) \citep{tsanas2009accurate} is to predict UPDRS scores from 42 people with early-stage Parkinson's disease. It contains 5875 instances and 20 features.
\end{itemize}

\paragraph{Algorithms Implementations.}\

The details regarding the implementation of our methods as well as the benchmarks are as follows:
\begin{itemize}
    \item NN+kNNRej (Algorithm \ref{alg:Calibrate}): We build a neural network  as the regressor $\hat{f}$. We then implement Algorithm \ref{alg:Calibrate} with radial basis function kernel $k(x_1,x_2)=\exp(\|x_1-x_2\|^2_2/\sigma)$ as the kernel choice, where the kernel length scale $\sigma$ is a hyperparameter tuning by the validation dataset. 
    \item SelNet: It contains one main body block, whose last layer is the representation layer to extract the data features, and three output heads for prediction, selection (rejection), and auxiliary prediction. The loss function is designed as the average over the original mean square loss of the auxiliary prediction head, the weighted mean square loss of the prediction head, where the weights are the output rejection confidence of the selection head and a penalty term to guarantee the overall rejection rate is smaller than a pre-determined rate $\gamma$. SelNet uses the hyperparameter $\alpha$ (with value $1/2$ in the original work and our implementation) to weight these two types of losses. Thus, SelNet jointly trains the predictor (prediction head) and the rejector (selection head), while utilizing the auxiliary prediction head to push the main body to learn all training instances to extract latent features and avoid overfitting.
    \item SelNet($\alpha=1$): same as the SelNet except $\alpha=1$. This means the auxiliary head's performance/loss is not included in the training loss and it only focuses on the to-be-accepted samples during the training and tends to ignore the to-be-rejected ones.
    \item NN+SelRej: It has the same architecture as the SelNet($\alpha=1$) except that the training has two stages: the first stage trains a neural network as a regressor $\hat{f}$ based on the training dataset and uses its main body (layers before the output layer) as the latent feature extractors for the second stage. In the second stage, we freeze this main body, add the prediction head and selection head as output layers and fine-tune them with the loss function from SelNet (with $\alpha=1$ since there is no auxiliary head). Thus, compared to SelNet($\alpha=1$), this two-stage training process forces the model to learn the latent features from all samples in the first stage.
    \item NN+LossRej: It is a variant of NN+SelRej: we still first train a neural network as a regressor and train another neural network, named loss rejector (LossRej), to predict the loss and as rejector. Specifically, the rejector network shares the same weights on the main body of the regressor network (as the feature extractor) and fine-tunes a linear output layer by validation data, which maps from latent features to loss.  This two-stage training process can again force both the regressor and rejector to learn from all samples. During the inference, if the deferral cost is given as $c$, then we reject the samples whose predicted losses are larger than $c$. And if a rejection rate budget $\gamma$ (and $c=0$) is given as in Table \ref{tab:apx_EXP2}, then we use the validation data to estimate a threshold of loss such that the samples with predicted losses larger than the threshold are rejected.
    \item NN+LogRej: It is the same as the NN+LossRej, except that the fine-tuned rejector network, named as logistic rejector (LogRej), directly predicts whether to reject samples: given validation data, we first decide the binary rejection labels based on the trained regressor (i.e., comparison with cost $c$ or choosing the top $\gamma$ fraction samples with largest losses), and fine-tune a linear layer with Sigmoid head as a binary classifier to decide the rejection. Thus, we make a logistic regression on the latent features and rejection labels.
        \item Triage: It first trains a neural network regressor. At training epoch $t$, it only uses the samples in the mini-batch with empirical mean square loss (based on the predictor fitted at epoch $t-1$) smaller than the deferral cost $c$, and thus to make the training process focus on the non-rejected samples. To improve the robustness of training, when the number of used samples from the mini-batch is smaller than $32$, it will use the first $32$ samples with the smallest empirical mean square losses. After training the predictor, it will train a neural network binary classifier as the rejector on the validation dataset, where the sample is labeled as positive if its empirical mean square loss (w.r.t. the fitted predictor) is less than $c$ and negative otherwise.  We highlight that the Triage method is the only algorithm among implemented algorithms that utilizes a subset of training data for training the predictor.
    \item kNN: It first trains a kNN-based predictor $\texttt{kNN}_y(x)$ for $\mathbb{E}\left[y\big \vert X=x\right]$ and trains another kNN-based predictor $\texttt{kNN}_l(x)$ for the loss $\mathbb{E}\left[l(\texttt{kNN}_y(x),Y)\big\vert X=x\right]$. For the testing sample $x$, the algorithm will reject it if $\texttt{kNN}_l(x)>c$ and will accept otherwise.
\end{itemize}

\paragraph{Architectures and Hyperparameters tuning.}\

The details regarding the architectures and hyperparameters tuning of our methods as well as the benchmarks are as follows. For benchmark algorithms, their architectures and hyperparameter tuning processes are set to be identical to the original works if applicable. 

\begin{itemize}
    \item NN+kNNRej: The neural network architecture for the regressor has one hidden layer activated by ReLU with 64 neurons and one following fully-connected linearly activated neuron. The kernel length scale $\sigma$ for the Gaussian kernel is selected to minimize the loss of validation data among $\sigma\in \{ 10^j: j=-3,-2,-1,0,1,2,3\}$.
    \item Triage: The neural network architecture for the predictor/regressor has one hidden layer activated by ReLU with 64 neurons and one following fully-connected  linearly activated neuron. The rejector also has one hidden layer activated by ReLU with 64 neurons and a following one Sigmoid-activated neuron. We set the training dataset as $\mathcal{D}_{\text{val}}$ due to the limited number of samples in most datasets. 
    \item SelNet/SelNet($\alpha=1$): The neural network architecture is identical to the original work for the regression task: the main body block has one hidden layer activated by ReLU with 64 neurons. Both prediction and auxiliary heads are fully connected with one linearly activated neuron. The selection head has one hidden layer activated by ReLu with 16 neurons and a following one Sigmoid-activated neuron. 
    If the cost $c$ is non-zero, we select the rejection rate $\gamma$ as a hyperparameter of the SelNet and use the validation data set to test the validation \texttt{RwR} loss for $\gamma \in \{0,0.2,0.4,0.6,0.8,1\}$ and select the $\gamma$ with the minimized validation loss. All other hyperparameters are identical to the original work. If the cost $c=0$ but a rejection rate budget is given as in Table \ref{tab:apx_EXP2},  as in the original work,  we use validation data to calibrate the threshold of rejection from $0,0.1,...,1$ to make the empirical rejection rate close to the given budget.
    \item NN+SelRej: The regressor network has the same architecture as in NN+kNNRej and the fine-tuned output layer (prediction and selection head) is identical to the SelNet. The dataset for the fine-tuning step is the same as the training dataset for the regressor network. We also use the same hyperparameter tuning process as the SelNet.
    \item NN+LossRej: The regressor network has the same architecture as in NN+kNNRej and the loss rejector network's head has one hidden layer activated by ReLu with 16 neurons with a linear output layer. The dataset for the fine-tuning step is the same as the training dataset for the regressor network. When making an inference for rejecting, the predicted probability from the rejector $>0.5$ will be rejected.
    \item NN+LogRej: The regressor network has the same architecture as in NN+kNNRej and the logistic rejector network's head has a linear layer that maps the latent features to a scalar, following a Sigmoid output layer. The dataset for the fine-tuning step is the same as the training dataset for the regressor network.
    \item kNN: We use the same hyperparameter turning processes to choose the number of neighbors $k$ as the original work, where we employ the 10-fold cross-validation to select the parameter $k \in \{5, 10, 15, 20, 30, 50, 70, 100, 150\}$ for two kNN models respectively.
\end{itemize}

All the neural network predictors/regressors' training/fine-tunings are optimized by the ADAM algorithm with a learning rate of $5\times 10^{-4}$ and weight decay of $1\times 10^{-4}$, and mini-batch size as 256 with 800 training epochs, which are all identical to the original SelNet setup \citep{geifman2019selectivenet} for the regression task. In addition, the neural network rejector in Triage is optimized by the ADAM algorithm with a learning rate of $1\times 10^{-3}$, and mini-batch size as 128 with 40 training epochs, which is identical to the setup in the posted code of the Triage algorithm \citep{okati2021differentiable}.

\section{Extension to Classification}
\label{appd:ext}

In this part, we consider binary classification with rejection under the weak realizability condition and show a parallel statement as Proposition \ref{prop:weak_real}. 

The setting of classification with rejection is the same as regression with rejection when we set the target space to be $\mathcal{Y}=\{0,1\}$. Specifically, this classification problem also encompasses two components: 
\begin{itemize}
    \item[(i)]
    a classifier $f:\mathcal{X}\rightarrow\{0,1\}$, which predicts the label from the feature

    \item[(ii)]
    a rejector $r:\mathcal{X}\rightarrow\{0,1\}$, which decides whether to predict with the predictor or abstain from the prediction to human.
\end{itemize}
Similarly, we also denote $\mathcal{F}$ be the set of candidate classifiers, and the goal is to find a pair of classifier and rejector to minimize the following expected \texttt{RwR} loss, which, in this case, is equivalent to 0-1 loss
\begin{align*}
    L_{\texttt{RWR}}
    =
    \mathbb{E}\left[
        r(X) \cdot I\{f(X)\not=Y\}
        +
        (1-r(X)) \cdot c
    \right].
\end{align*}

In this case, Definition \ref{def:wr} cannot be satisfied in general. The reason is that the conditional expectation function $\bar{f}=\mathbb{E}[Y|X=x]$ might be a fractional number, and thus it cannot be a classifier. Therefore, we first redefine the weak realizability for classification:
\begin{definition}[Weak realizability for classification]
    The joint distribution of the feature and label pair $(X,Y)$ and the function class $\mathcal{F}$ satisfy \textit{weak realizability} if the classifier is consistent with the conditional expectation function $\bar{f}$ is in the function class. That is, the following classifier is in the function class $\mathcal{F}$.
    \begin{align*}
        \tilde{f}(x)
        =
        \begin{cases}
            1, &\text{if $\bar{f}(x)=\mathbb{E}[Y|X=x]\geq0.5$,}\\
            0, &\text{otherwise.}
        \end{cases}
    \end{align*}

\end{definition}

Then, similar to Proposition \ref{prop:weak_real} for regression with rejection, we have a similar result for classification with rejection.
\begin{proposition}
\label{prop:weak_real_class}
If the weak realizability condition for classification holds, then $\tilde{f}(\cdot)$ minimizes the expected \texttt{RwR} loss that
\begin{align*}
\tilde{f}(\cdot) \in\argmin_{f\in\mathcal{F}}L_{\mathtt{RwR}}(f,r)
    \end{align*}
for any measurable rejector function $r(\cdot).$
\end{proposition}
\begin{proof}
    Similar to the proof of Proposition \ref{prop:weak_real}, we directly minimize the expected \texttt{RwR} loss for any fixed feature $X\in\mathcal{X}$. For any fixed $X\in\mathcal{X}$, we have
    \begin{align*}
        \mathbb{E}_Y\left[l_{\texttt{RwR}}(f,r)|X\right]
        =
        \mathbb{E}_Y\left[(Y-f(X))^2|X\right]\cdot r(X).
    \end{align*}
    If $r(X)=0$, the corresponding \texttt{RwR} loss is $0$ given the feature $X$, and thus, all predictors minimize the \texttt{RwR} loss at that feature. If $r(X)=1$, we have
    \begin{align*}
        \mathbb{E}_Y\left[(Y-f(X))^2|X\right]
        =
        \mathbb{E}[(Y-\bar{f}(X))^2|X]+\mathbb{E}[(f(X)-\bar{f}(X))^2|X],
    \end{align*}
    where the first term is independent of the choice of the predictor, and the second term is minimized at $f=\tilde{f}$. We then finish the proof.     
\end{proof}

\section{Auxiliary Lemmas}

We introduce a lemma related to the uniform laws of large numbers, which is a useful tool when proving consistency.

\begin{lemma}[Uniform Laws of Large Numbers]
	\label{lem:ulln} Let $\mathcal{F}=\{f_{\theta},\theta\in\Theta\}$ be a collection of measurable functions defined on $\mathcal{X}$ indexed by a bounded subset $\Theta\subseteq\mathbb{R}^d$. Suppose that there exists a constant $M$ such that $\left\vert  f_{\theta_1}(X)- f_{\theta_2}(X) \right\vert\leq M\| \theta_1-\theta_2\|_2$. Then, 
    \begin{align*}
        \sup_{f\in\mathcal{F}}\left|\frac{1}{n}\sum\limits_{i=1}^{n} f(X_i)
        - \mathbb{E}[f(X)]
        \right|
        \rightarrow
        0
    \end{align*}
    almost surely as $n\rightarrow\infty$, where the expectation is taken with respect to i.i.d. samples $\{X_i\}_{i=1}^{\infty}$.
\end{lemma}
\begin{proof}
    We refer to 19.5 Theorem (Glivenko-Cantelli) and 19.7 in \cite{van2000asymptotic}.
\end{proof}

\section{Proofs of Section \ref{sec:wr}}
\label{appd:pf2}

\subsection{Proof of Proposition \ref{prop:subopt}}

\begin{proof}
Part (a): we construct a regressor $f_0$ via
\[
f_0(x) \coloneqq \begin{cases}
\mathbb{E}[Y|X=x] + 2\sqrt{c}, \quad & \text{if } \mathbb{E}[(Y - \mathbb{E}[Y|X=x])^2 |X=x] \leq c;\\

\mathbb{E}[Y|X=x], \quad & \text{if } \mathbb{E}[(Y - \mathbb{E}[Y|X=x])^2 |X=x] > c.
\end{cases}
\]
$f_0$ performs poorly on every instance of $X=x$ such that the expected conditional squared loss is at least $4c$ for those low variance cases and at least $c$ for those high variance cases. Thus, we can construct its optimal rejector $r_0$ to be $r_0 \coloneqq 0$, inducing an \texttt{RwR} loss of $c$.

For any regressor $f^\prime_0$ that slightly deviates from $f_0$ by at most $\sqrt{c}$, it still performs poorly on every instance of $X=x$ such that the expected conditional squared loss is at least $c$ no matter which rejector it uses. Thus, $(f_0, r_0)$ is locally optimal in a neighborhood of radius $\sqrt{c}$. But apparently, it is not globally optimal.

Part (b): consider an instance of $X=x$ with the conditional variance of $Y$ smaller than $c$. By the continuous assumption, there is a neighborhood of $x$ (denoted by $U_1$) such that the conditional variance of $Y$ is permanently smaller than $c$ on $U_1$. W.l.o.g. we assume the probability of $X$ falling into $U_1$ is lower bounded by $p_1 > 0$. Then we can construct a regressor $f_1$ as
\[
f_1(x) \coloneqq \begin{cases}
\mathbb{E}[Y|X=x] + 2\sqrt{c}, \quad & \text{if } x \in U_1;\\

\mathbb{E}[Y|X=x], \quad & \text{otherwise}.
\end{cases}
\]
We denote the optimal rejector induced by $f_1$ as $r_1$. Then the pair $(f_1, r_1)$ must be optimal w.r.t. the rejector argument. As for the regressor argument, notice that $r_1(x) = 0$ for any $x \in U_1$, which means that any change for $f_1(x), x \in U_1$ does not affect the \texttt{RwR} loss. Combining this with the fact that $f_1$ is optimal outside $U_1$, we conclude that $(f_1, r_1)$ is optimal w.r.t. the regressor argument. But it is not globally optimal.
\end{proof}

\subsection{Proof of Proposition \ref{prop:weak_real}}

\begin{proof}
To prove the statement, we directly minimize the expected \texttt{RwR} loss for any fixed feature $X\in\mathcal{X}$ and rejector $r$. We first show that for any fixed feature $X$ and any fixed rejector $r$, the conditional expectation $\bar{f}(X)=\mathbb{E}[Y|X]$ minimizes the conditional \texttt{RwR} loss $\mathbb{E}_{Y}[(Y-f(X))^2\cdot r(X)|X]$ with respect to the predictors $f\in\mathcal{F}$. To see this, we first notice when $r(X)=0$, the conditional \texttt{RwR} is constant and all predictors minimize this loss; when $r(X)=1$, 
\begin{align*}
\mathbb{E}_{Y}[(Y-f(X))^2\cdot r(X)|X] &= \mathbb{E}_{Y}[(Y-f(X))^2|X]\\
&= \mathbb{E}_{Y}[(Y-\bar{f}(X))^2|X] + (\bar{f}(X) - f(X))^2,
\end{align*}
which is minimized by $\bar{f}(X)=\mathbb{E}[Y|X]$.

Then, since the \texttt{RwR} loss is minimized pointwise by $\bar{f}(\cdot)$, we have that when the weak realizability condition holds, i.e., $\bar{f}\in\mathcal{F}$, $\bar{f}(\cdot)$ also minimizes the expected \texttt{RwR} loss $L_{\texttt{RwR}}$.
\end{proof}

\subsection{Proof of Proposition \ref{prop:consistency}}
\begin{proof}
With a fixed rejector $r$, one can easily verify
\begin{align*}
L_{\mathtt{RwR}}(f_n, r) - L_{\mathtt{RwR}}(f^*, r) & = \E_X[r(X) \cdot (\E_Y[(f_n(X) - Y)^2|X] - \E_Y[(f^*(X)- Y)^2|X]) + (1-r(X)) \cdot (c-c)]\\
& = \E_X[r(X) \cdot (f_n(X) - f^*(X))^2]\\
& \leq \E_X[(f_n(X) - f^*(X))^2] \\
& = \E[(f_n(X) - Y)^2] - \E[(f^*(X) - Y)^2],
\end{align*}
where the second equality holds by bias-variance decomposition, the inequality by $0\leq r(X) \leq 1$, and the last equality by bias-variance decomposition again.
Since $\E[(f_n(X) - Y)^2] - \E[(f^*(X) - Y)^2]$ tends to zero in probability (or almost surely) as $n$ tends to infinity, the proof is complete.

For the reverse part, one just needs to set $r(X) \equiv 1$ to verify the conclusion.
\end{proof}

\subsection{Proof of Example \ref{eg:parametrized_consistency}}
\begin{proof}
We first show that $\mathbb{E}[(Y-f_{\theta_n}(X))^2]$ converges to $\inf_{\theta\in\Theta}\mathbb{E}[(Y-f_{\theta}(X))^2]$ almost surely as $n$ goes to infinity. This part is an application of the uniform laws of large numbers: by Lemma \ref{lem:ulln}, we have $\mathbb{E}[(Y-f_{\theta_n}(X))^2]$ converges to $\inf_{\theta\in\Theta}\mathbb{E}[(Y-f_{\theta}(X))^2]$ almost surely as $n\rightarrow \infty$.

Then, we show that $\inf_{\theta\in\Theta}\mathbb{E}[(Y-f_{\theta}(X))^2]=\mathbb{E}[(Y-f^*(X))^2]$. In fact, this is a direct consequence of the weak realizability condition of $\mathcal{F}$, which states that $\bar{f} \in \mathcal{F}$ and $\bar{f}$ minimizes the conditional risk pointwise (see the Proof of Proposition \ref{prop:weak_real}).

We now have $\mathbb{E}[(Y-f_{\theta_n}(X))^2] \rightarrow \mathbb{E}[(Y-f^*(X))^2]$ almost surely as $n \rightarrow \infty$. Applying Proposition \ref{prop:consistency} yields the desired result.

\end{proof}

\subsection{Consistency of the regressor-rejector pair}

By Proposition \ref{prop:weak_real}, the optimal rejector for any regressor $f$ is $r_f$. We can then formalize the consistency of any rejector w.r.t. the convergence to $r_f$. Combining a consistent regressor with another consistent rejector (where we defer the learning of rejectors to later sections) leads to a consistent regressor-rejector pair.

\begin{proposition}
\label{prop:pair_consistency}
Suppose we have a consistent regressor $\{f_n\}_{n=1}^{\infty}$ and a consistent rejector $\{r_n\}_{n=1}^{\infty}$, i.e.,
\[\E[(f_n(X) - f^*(X))^2] \rightarrow 0\]
and
\[\E[(r_n(X) - r_{f_n}(X))^2] \rightarrow 0\]
in probability (or almost surely) as $n \rightarrow \infty$. Assume that all $f_n(\mathcal{X})$'s and $\mathcal{Y}$ are uniformly bounded.
Then we have a consistent regressor-rejector pair $\{(f_n, r_n)\}_{n=1}^{\infty}$, say
\[L_{\mathtt{RwR}}(f_n, r_n) \rightarrow L_{\mathtt{RwR}}(f^*, r^*)\]
in probability (or almost surely) as $n \rightarrow \infty$.
\end{proposition}

\subsection{Proof of Proposition \ref{prop:pair_consistency}}
\begin{proof}
Recall that we have defined $r_f(x) = \mathbbm{1}\{\E[(f(X) - Y)^2|X=x] \leq c\}$. We first prove that $L_{\mathtt{RwR}}(f_n ,r_n)$ tends to $L_{\mathtt{RwR}}(f_n, r_{f_n})$. This is due to
\begin{align*}
0 \leq L_{\mathtt{RwR}}(f_n, r_n) - L_{\mathtt{RwR}}(f_n, r_{f_n}) & = \E_X[\E[(f_n(X)-Y)^2|X] \cdot (r_n(X) - r_{f_n}(X)) + c \cdot (r_{f_n}(X) - r_n(X))] \\
& \leq \E_X[(\E[(f_n(X)-Y)^2|X]+c) \cdot |r_n(X) - r_{f_n}(X)|]\\
& = \E_X[(\E[(f_n(X)-Y)^2|X]+c) \cdot (r_n(X) - r_{f_n}(X))^2]\\
& \leq \E_X[B (r_n(X) - r_{f_n}(X))^2] \rightarrow 0,
\end{align*}
as $n \rightarrow 0$ in probability (or almost surely), where the first inequality is due to the optimality of $r_{f_n}$ w.r.t. $f_n$, the second inequality due to the non-negativity of each part, the last equality due to the fact that $|r_n(X) - r_{f_n}(X)|$ is boolean, and the last inequality by the uniform boundedness of $f_n(\mathcal{X})$ and $\mathcal{Y}$ by some constant $B$.

Since
\begin{align*}
L_{\mathtt{RwR}}(f_n, r_{f_n}) & = \E_X[\E[(f_n(X)-Y)^2 |X] \wedge c], \\
L_{\mathtt{RwR}}(f^*, r^*) & = \E_X[\E[(f^*(X)-Y)^2 |X] \wedge c],
\end{align*}
we have
\begin{align*}
0 \leq L_{\mathtt{RwR}}(f_n, r_{f_n}) - L_{\mathtt{RwR}}(f^*, r^*) & = \E_X[\E[(f_n(X)-Y)^2 |X] \wedge c - \E[(f^*(X)-Y)^2 |X] \wedge c]\\
& \leq \E_X[\E[(f_n(X)-Y)^2 |X] - \E[(f^*(X)-Y)^2 |X]]\\
& = \E[(f_n(X) - Y)^2] - \E[(f^*(X) - Y)^2]\\
& = \E[(f_n(X) - f^*(X))^2] \rightarrow 0,
\end{align*}
as $n \rightarrow 0$ in probability (or almost surely), where the first inequality holds by the optimality of $(f^*, r^*)$, the second inequality by the fact that $a \wedge c - b \wedge c \leq a - b$ for any $ a \geq b$, and the last equality by the bias-variance decomposition.
\end{proof}

\section{Proof of Section \ref{sec:lbwr}}
\label{appd:pf3}

\subsection{Proof of Proposition \ref{prop:trun}}
\begin{proof}
	\begin{itemize}
		\item[(a)]
		We first show that the expected \texttt{RwR} loss is bounded by the truncated loss from the below. To see this,
		\begin{align*}
			\mathbb{E}_{Y} [l_{\texttt{RwR}}(f,r;(x,Y))|X]
			&=
			\mathbb{E}_{Y}[(Y-f(X))^2 |X]\cdot r(X) + c\cdot(1-r(X))\\
			&\geq
			\left( \mathbb{E}_{Y}[(Y-f(X))^2|X ]\land c\right)\cdot r(X)+ \left( \mathbb{E}_{Y}[(Y-f(X))^2|X ]\land c\right)\cdot (1-r(x))\\
			&=
			\left( \mathbb{E}_{Y}[(Y-f(X))^2|X] \land c\right),
		\end{align*}
		where this equality comes from the definition of the \texttt{RwR} loss, the second line from the monotonicity of the $\min$ function, and the last line comes from the direct calculation. The equality holds if $r(X) = \mathbbm{1}\{\E_Y[(f(X) - Y)^2 | X] \leq c\}$. Then, taking expectation with respect to the feature $X$, we finish the proof of Part (a).
		
		\item[(b)]
	   We have
        \allowdisplaybreaks
        \begin{align*}
        & \phantom{=} L_{\mathtt{RwR}}(f, r_{\hat{R}}) - \tilde{L}(f) \\
        & = \E_X\Big[\E_Y\big[(f(X) - Y)^2|X\big] \cdot r_{\hat{R}}(X) + c \cdot (1-r_{\hat{R}}(X)) - \E_Y\big[(f(X) - Y)^2 | X\big]\wedge c\Big]\\
        & = \E_X\bigg[\Big(\E_Y\big[(f(X) - Y)^2 | X\big] - c\Big) \cdot \mathbbm{1}\big\{r_{\hat{R}}(X) = 1 \text{ and } \E_Y[(f(X) - Y)^2 | X] > c\big\} \\
        & \phantom{===} + \Big(c - \E_Y\big[(f(X) - Y)^2 | X\big]\Big) \cdot \mathbbm{1}\big\{r_{\hat{R}}(X) = 0 \text{ and } \E_Y[(f(X) - Y)^2 | X] \leq c\big\}\bigg]\\
        & = \E_X\Big[\big(R(f, X) - c\big) \cdot \mathbbm{1}\big\{\hat{R}(f, X) \leq c \text{ and } R(f, X) > c\big\} \\
        & \phantom{===}+ \big(c - R(f, X)\big) \cdot \mathbbm{1}\big\{\hat{R}(f, X) > c \text{ and } R(f, X) \leq c\big\}\Big] \\
        & \leq \E_X\Big[\big(R(f, X) - \hat{R}(f, X)\big) \cdot \mathbbm{1}\big\{\hat{R}(f, X) \leq c \text{ and } R(f, X) > c\big\}\\
        & \phantom{===}+ \big(\hat{R}(f, X) - R(f, X)\big) \cdot \mathbbm{1}\big\{\hat{R}(f, X) > c \text{ and } R(f, X) \leq c\big\}\Big] \\
        & \leq \E_X\Big[\big\vert \hat{R}(f, X) - R(f, X) \big\vert\Big],
        \end{align*}
        where the first two equalities hold by definition of $L_{\mathtt{RwR}}$ and $\tilde{L}$, the third equality by the definition of $R$ and $\hat{R}$, the first inequality by checking the indicator functions' conditions, and the second inequality by the fact that those two indicator functions cannot be both non-zero simultaneously.
	\end{itemize}
\end{proof}

\subsection{Proof of Proposition \ref{prop:cali}}

\begin{proof}
we first show that both the expected squared loss and the truncated loss can be minimized by the conditional expectation function $\bar{f}(X)$. For any predictor, its squared loss can be re-written as follows:
\begin{align}
	\label{eq:l2}
	L_2(f) &= \mathbb{E}[(Y-f(X))^2]\nonumber \\
	&=
	\mathbb{E}[(Y-\bar{f}(X)+\bar{f}(X)-f(X))^2]\\
	&=
	\mathbb{E}[(Y-\bar{f}(X))^2]+\mathbb{E}[(\bar{f}(X)-f(X))^2]+2\mathbb{E}[(Y-\bar{f}(X))(\bar{f}(X)-f(X))]\nonumber\\
	&=
	\mathbb{E}[(Y-\bar{f}(X))^2]+\mathbb{E}[(\bar{f}(X)-f(X))^2], \nonumber
\end{align}
where the first line comes from the definition of the squared loss, the second and the third line comes from direct calculation, and the last line from the equality that $\mathbb{E}[(Y-\bar{f}(X))(\bar{f}(X)-f(X))]=0$ obtained by the definition of $\bar{f}(X)$. In the last line of \eqref{eq:l2}, the first term is independent of the chosen predictor. The second term is non-negative, and it is zero only when $\bar{f}(X)=f(X)$ with probability one. Then, we can see that for any predictor $f\not=\bar{f}$, the corresponding squared loss is no less than the squared loss of $\bar{f}$. Thus, we have the squared loss is minimized by $\bar{f}$. For the truncated loss, the proof is similar. We have
\begin{align}
	\label{eq:l_trun}
	\tilde{L}(f) &= \mathbb{E}_X[\mathbb{E}_Y[(Y-f(X))^2|X]\land c]\nonumber \\
	&=
	\mathbb{E}_X\left[\left(\mathbb{E}_Y[(Y-\bar{f}(X))^2|X]+\mathbb{E}_Y[(\bar{f}(X)-f(X))^2|X]\right)\land c\right]\\
	&\geq
	\mathbb{E}_X\left[\mathbb{E}_Y[(Y-\bar{f}(X))^2|X]\land c\right] \nonumber,
\end{align}
where the first line comes from the definition of the truncated loss, the second line comes from similar steps as in \eqref{eq:l2}, and the last line comes from the non-negativity of $\mathbb{E}_Y[(\bar{f}(X)-f(X))^2|X]$. Thus, similarly, we can see that the truncated loss is also minimized by $\bar{f}$. Then, recall that \[
    \tilde{L}^* \coloneqq \min_{f} \tilde{L}(f),\ L_2^* \coloneqq \min_{f} L_2(f),
\]
where the minimums are taken over all measurable functions. Thus, we have showed 
        \begin{align}
            \label{cond:cali_cons}
            \tilde{L}(\bar{f})=\tilde{L}^*,\ L_2(\bar{f}) = L_2^*.
        \end{align}

Next, we show the calibration condition \eqref{ieq:cali_func}. For any predictor $f$, we have
\begin{align}
	\label{ieq:ex_l_trun}
	\tilde{L}(f) - \tilde{L}^*
	&=
	\tilde{L}(f) - \tilde{L}(\bar{f})\nonumber\\
	&=
	\mathbb{E}_X\left[\left(\mathbb{E}_Y[(Y-\bar{f}(X))^2|X]+\mathbb{E}_Y[(\bar{f}(X)-f(X))^2|X]\right)\land c
	-
	\mathbb{E}_Y[(Y-\bar{f}(X))^2|X]\land c\right] \nonumber\\
	&\leq
	\mathbb{E}_X\left[\mathbb{E}_Y[(\bar{f}(X)-f(X))^2|X]\land c\right]\\
	&\leq
	\mathbb{E}\left[(\bar{f}(X)-f(X))^2\right]\nonumber
\end{align}
where the first line is obtained by \eqref{cond:cali_cons}, the second and last lines come from \eqref{eq:l2} and \eqref{eq:l_trun}, the third line comes from the inequality that 
\[
	\min(x,c)+\min(y,c)\geq\min(x+y,c)
\] 
for any non-negative numbers $x,y$, and the forth line comes from $x\geq \min(x,c)$ for all $x$.

Finally, to show the calibration condition, it is sufficient to show that the last line of \eqref{ieq:ex_l_trun} equals to $L_2(f)-L_2^*$ for any predictor $f$. This can be obtained by \eqref{eq:l2},
\begin{align*}
	L_2(f)-L_2^*
	&=
	L_2(f)-L_2(\bar{f}) \nonumber \\
	&=
	\mathbb{E}[(Y-\bar{f}(X))^2]+\mathbb{E}[(\bar{f}(X)-f(X))^2] - \mathbb{E}[(Y-\bar{f}(X))^2]\\
	&=
	\mathbb{E}[(\bar{f}(X)-f(X))^2] \nonumber,
\end{align*}
where the first line comes from \eqref{cond:cali_cons}, the second line comes from \eqref{eq:l2}, and the last line comes from cancelling  $\mathbb{E}[(Y-\bar{f}(X))^2]$.

\end{proof}

\subsection{Proof of Theorem \ref{thm:gen_decompose}}

\begin{proof}
	The statement of Theorem \ref{thm:gen_decompose} can be obtained directly from Proposition \ref{prop:cali} and Part (b) of Proposition \ref{prop:trun}. Specifically, we have
\begin{align*}
	L_{\mathtt{RwR}}(\hat{f}, r_{\hat{R}})
	-
	L^*_{\mathtt{RwR}}
	&\leq
	\tilde{L}(\hat{f}) - \tilde{L}^* + 
	\E\big[\vert \hat{R}(\hat{f}, X) - R(\hat{f}, X)\vert\big]\\
	&\leq L_2(\hat{f}) - L_2^*
		+\E\big[\vert \hat{R}(\hat{f}, X) - R(\hat{f}, X)\vert\big]\\
    &= \E\Big[\big(\hat{f}(X) - f^*(X)\big)^2\Big]
		+\E\big[\vert \hat{R}(\hat{f}, X) - R(\hat{f}, X)\vert\big],
\end{align*}
where the first line is obtained by Part (b) of Proposition \ref{prop:trun}, the second line from Proposition \ref{prop:cali}, and the last line from the bias-variance decomposition.

\end{proof}

\end{document}